\documentclass[a4paper]{article}
\usepackage{amsmath,amssymb,mathtools,bbm,pifont,tipa,booktabs,caption,subcaption,url,natbib}
\usepackage{multirow}
\usepackage[multiple]{footmisc}
\usepackage{url}
\usepackage{hyperref}
\usepackage[hyphenbreaks]{breakurl}
\usepackage[textwidth=6.5in,textheight=710pt]{geometry}
\usepackage{authblk}

\newcommand{\checkMark}{\ding{51}} 
\newcommand{\supcheckMark}{\textsuperscript{\checkMark}}
\newcommand{\escapedURL}[1]{\href{#1}{\detokenize{#1}}}

\title{Unsupervised Classification of English Words \\ Based on Phonological Information: \\ Discovery of Germanic and Latinate Clusters}

\author[1,2]{Takashi Morita\thanks{tmorita@alum.mit.edu}}
\affil[1]{Academy of Emerging Sciences, Chubu University}
\affil[2]{Institute for Advanced Research, Nagoya University}
\author[3,4,5]{Timothy J. O'Donnell\thanks{timothy.odonnell@mcgill.ca}}
\affil[3]{Department of Linguistics, McGill University}
\affil[4]{Canada CIFAR AI Chair, Mila}
\affil[5]{Mila — Quebec AI Institute}

\begin{document}
    \maketitle
    \begin{abstract}
        Cross-linguistically, native words and loanwords follow different phonological rules. In English, for example, words of Germanic and Latinate origin exhibit different stress patterns, and a certain syntactic structure---double-object datives---is predominantly associated with Germanic verbs rather than Latinate verbs. From the perspective of language acquisition, however, such etymology-based generalizations raise learnability concerns, since the historical origins of words are presumably inaccessible information for general language learners. In this study, we present computational evidence indicating that the Germanic-Latinate distinction in the English lexicon is learnable from the phonotactic information of individual words. Specifically, we performed an unsupervised clustering on corpus-extracted words, and the resulting word clusters largely aligned with the etymological distinction. The model-discovered clusters also recovered various linguistic generalizations documented in the previous literature regarding the corresponding etymological classes. Moreover, our model also uncovered previously unrecognized features of the quasi-etymological clusters. Taken together with prior results from Japanese, our findings indicate that the proposed method provides a general, cross-linguistic approach to discovering etymological structure from phonotactic cues in the lexicon.
    \end{abstract}



\section{Introduction}
Discovering appropriate groups of observations without access to correct answers (i.e., unsupervised class learning/clustering) is a fundamental challenge in the computational modeling of language acquisition. For example, a plausible model of spoken-language learners must be able to identify the vowel and consonant inventories of the target language solely from the acoustic properties of speech inputs (without reference to text transcriptions, unlike industrial speech recognition systems). This phonetic learning is particularly challenging due to the considerable individual and contextual variations in the acoustic data \citep{Vallabha+07,Feldman+09,Dunbar+17_ZeroSpeech2017}.

In addition to categorizing individual speech sounds (phonemes), language learners must also acquire knowledge of their discrete sequential patterns (phonology). This phonological learning also involves a classification task, since the lexicon of a single language comprises multiple groups of words that adhere to different phonological rules and constraints. For example, nouns and verbs are often governed by separate phonological rules/constraints in various languages \citep{GoodenoughSuigita80,KellyBock88,Bobaljik97,Meyers97,Smith99,Smith16}. Likewise, semantically distinguished classes of words may also exhibit unique phonological patterns, differing from the rest of the lexicon within the same language; for instance, onomatopoeic expressions (ideophones) in Japanese are formed through the repetition of a bimoraic morpheme \citep[e.g., \lbrack ki{\textfishhookr}a-ki{\textfishhookr}a \rbrack, meaning ``shining'';][]{ItoMester95JPhon,ItoMester99}, and similar exceptionalities of this word class have been documented across languages \citep[see][for a review]{Dingemanse12}. 

Similarly, words of different etymological origins exhibit different phonological patterns. For example, English words are typically categorized according to their Germanic vs. Latinate origins, and this distinction correlates with two different stress patterns found in verbs \citep{Grimshaw85,GrimshawPrince86}. Etymological classification also provides a crucial framework for analyzing English morphology, as Latinate suffixes predominantly attach to Latinate roots, thereby maintaining etymological consistency throughout entire words \citep{Anshen+86,Fabb88,ODonnell15}. Comparable etymology-based generalizations have been reported in other languages as well \citep[see the next section for details;][]{Trubetzkoy39,FriesPike49,Lees61,Postal69,Zimmer69,Chung83}.
Furthermore, experimental studies indicate that native speakers possess knowledge of such etymological distinctions, supporting their psychological reality beyond mere statistical observation \citep{Gropen+89,MoretonAmano99}.

However, modeling the acquisition of such etymology-dependent patterns poses exceptional challenges compared to other class-dependent patterns. The etymological origins of words are not directly observable to general language learners, unlike the abundant syntactic and semantic information embedded in word-sequence data \citep{Mikolov+13_word2vec,Mikolov+13_NIPS,Radford+19,Brown+20,Ouyang+22}. A recent study by \citet[see also \citealp{Morita_thesis}]{MoritaODonnell_sublexical_phonotactics} introduced a computational framework to investigate the learnability of etymological distinctions in the absence of explicit cues (i.e., without supervision). Case-studying Japanese, they demonstrated that etymological word classes of the language can be inferred solely from phonological information. Specifically, they performed unsupervised clustering on existing Japanese words, represented as strings of phonetic symbols, and their learning model identified word clusters that were significantly well-aligned with the etymologically defined classes. Moreover, the model also recovered the etymology-based phonological generalizations previously described in the literature. These findings offer an empirical justification for linguistic theories that posit etymological distinctions in the mental grammar, by showing that such distinctions need not be acquired through direct access to words' historical origins (i.e., with supervision), but can instead be inferred from observable phonotactic information.

In the present study, we apply the same clustering algorithm to English words, and demonstrate that the distinction between Germanic and Latinate origins is also learnable from sequential patterns of phonetic symbols appearing in the words (i.e., segmental phonotactics).
Our contributions can be summarized as follows.
\begin{itemize}
\item We present empirical evidence for the unsupervised learnability of the Germanic and Latinate word clusters based solely on phonological information, or segmental phonotactics in particular.
\item We demonstrate that the identified word clusters recover various linguistic properties of Germanic and Latinate words as proposed in the previous literature.
\item We highlight several hitherto unnoticed linguistic properties of the discovered word clusters, which can guide future experimental studies.
\item In conjunction with the findings from the previous study on Japanese \citep{MoritaODonnell_sublexical_phonotactics}, our present study supports the cross-linguistic validity of the proposed learning framework.
\end{itemize}

The remainder of this paper is organized as follows. 
In the next section, we commence with a review of related studies concerning etymology-based generalizations of morpho-phonological patterns in English and other languages. 
We then outline our model for learning etymological classes (or their proxies) exclusively from phonotactic information, alongside an explanation of the dataset employed for the learning simulation. 
As an outcome of the simulation under these settings,
we first report the basic results of the word clustering, including the degree of alignment between the model-detected clusters and the ground-truth etymology.
We then examine the linguistic properties of the identified clusters, demonstrating how they recover etymology-based generalizations previously noted in the literature. 
The final section provides a comprehensive discussion of our findings, outlines potential avenues for future research, and offers concluding remarks.

\section{Background} \label{sec:related-works}
\subsection{Cross-Linguistic Ubiquity of Etymology-Based Phonology}
Etymologically-defined lexical subclasses have been extensively documented for various languages, most typically distinguishing between native words and loanwords \citep[see][for reviews]{ItoMester95JPhon}. For example:
\begin{itemize}
\item In Chamorro, mid vowels are absent in native words but present in Spanish loans \citep[][]{Chung83}.
\item In Mohawk, labial consonants {\lbrack m,b,p\rbrack} are found in French loans but not in native words \citep{Postal69}.
\item In Mazateco, postnasal stops are systematically voiced in native words but can be voiceless in loans \citep{FriesPike49}.
\item German native words do not start with {\lbrack s\rbrack}, whereas this constraint does not apply to loans \citep{Trubetzkoy39}.
\item In Turkish, high vowels are rounded after labial consonants in native words but not necessarily in loans \citep{Lees61,Zimmer69}.
\end{itemize}
Etymological distinctions are not necessarily binary. For instance, Japanese has ternary distinctions in its morpho-phonology; words are divided into native words, loanwords from Old Chinese, and more recent loans primarily from English \citep{ItoMester95lexicon,ItoMester95JPhon,ItoMester99,ItoMester03,ItoMester08,Fukazawa98,Fukazawa+98_lexicalstratification,MoretonAmano99,Ota04,GelbartKawahara07,Frellesvig10}.
Comparable non-binary distinctions have also been reported for other languages, including Russian \citep{Holden76}, Turkish \citep{Zimmer85}, and Qu\'{e}bec French \citep{ParadisLebel94,HsuJesney17,HsuJesney18}.

\subsection{Etymology-Based Generalizations in English}
Like other languages, English exhibits linguistic generalizations rooted in the Germanic-Latinate distinction. One such generalization involves the etymological consistency of morphemes within a word: Latinate bases tend to be suffixed with Latinate morphemes \citep{Anshen+86,Fabb88,ODonnell15}. Another well-known example is the difference in stress patterns: Germanic verbs bear initial stress, whereas the initial syllable of Latinate verbs is typically unstressed \citep{Grimshaw85,GrimshawPrince86}. In addition, velar stops [k] and [g] are subject to ``softening'' into [s] and [\textdyoghlig] respectively in Latinate words (e.g., \textit{electri\underline{c}}$\to$\textit{electri\underline{c}ity}), but not in Germanic ones \citep{ChomskyHalle68,Pierrehumbert06}.

The Germanic-Latinate distinction in English also predicts a difference in syntactic patterns of verbs. Most famously, Germanic and Latinate verbs differ in their compatibility with the \emph{double-object construction}. On the one hand, the dative argument of Germanic verbs (e.g., \textit{bring}) can be expressed either as a prepositional phrase or as an indirect object:
\begin{itemize}
\item {\checkMark}\textit{The technician brought the new device \underline{to the professor}.}\\*
        (prepositional construction)
\item {\checkMark}\textit{The technician brought \underline{the professor} the new device.}\\*
        (double-object construction)
\end{itemize}
On the other hand, Latinate verbs are generally considered to disallow the double-object construction \citep{Levin93}. For example, the dative argument of the verb \textit{deliver}---despite its semantic similarity to \textit{bring}---can only appear as a prepositional phrase, not as an indirect object:
\begin{itemize}
    \item {\checkMark}\textit{The technician delivered the new device \underline{to the professor}.}\\*
            (prepositional construction)
    \item *\textit{The technician delivered \underline{the professor} the new device.}\\*
            (double-object construction)
\end{itemize}
This syntactic contrast extends to newly coined words as well. \citet{Gropen+89} experimentally demonstrated that the double-object construction was judged more grammatically acceptable with quasi-Germanic (monosyllabic) nonce verbs than with quasi-Latinate (polysyllabic) nonce verbs.

As a caveat, there are exceptions to this etymology-based generalization. For instance, \textit{pass} is a Latinate verb that nonetheless permits the double-object construction:
\begin{itemize}
\item {\checkMark}\textit{The technician passed the new device \underline{to the professor}.}\\*
        (prepositional construction)
\item {\checkMark}\textit{The technician passed \underline{the professor} the new device.}\\*
        (double-object construction)
\end{itemize}
Interestingly, our learning model---introduced in the next section---``correctly misclassified'' such exceptions into the Germanic-like word cluster on the basis of their phonotactic Germanicity, thereby outperforming grammaticality predictions based on true etymological origin.

\section{Materials and Methods} \label{sec:materials-and-methods}

This section provides a high-level explanation of our learning model for inducing word clusters corresponding to etymological distinctions,
and the dataset used to train it.

\subsection{Overview of the Learning Model} \label{sec:clustering-method-overview}
We model the learning of etymological lexical classes in the framework of unsupervised word clustering. The learning model receives English words---represented as strings of phonetic symbols---as its inputs and groups them into an optimal number of clusters following a certain statistical policy (explained below). Most importantly, the model has no access to ground-truth etymological information (such as ``Germanic origin'' or ``Latinate origin'') during its learning process, making it \emph{unsupervised}. This setup more closely approximates human language acquisition than approaches rely on access to ground-truth etymology (i.e., supervised learning).

Our approach to word clustering is grounded in Bayesian inference \citep{Anderson90,Tenenbaum99thesis}: We define a prior probability of word partitions (i.e., learning hypotheses) as well as their likelihood with respect to the data. Then, the optimal clustering is determined by the maximization of the posterior probability, which is proportional to the product of the prior and the likelihood.

In the remainder of this section, we focus on a high-level explanation of the model components, abstracting away from mathematical details. Interested readers are referred to \citet{MoritaODonnell_sublexical_phonotactics}. In addition, the Python code used for this study is publicly available in the public repository.%
\footnote{
    \escapedURL{https://github.com/tkc-morita/variational_inference_DP_mix_HDP_topic_ngram}.}
The values of hyperparameters are reported in Appendix~\ref{sec:hyperparameters}.

\subsubsection{Prior}

We adopt a non-parametric prior distribution on cluster assignments known as the \textit{Dirichlet process} \citep[DP;][]{Ferguson73,Antoniak74,Sethuraman94}. The DP prior prioritizes grouping words into the fewest possible clusters; it assigns a geometrically smaller probability (in expectation) to assignments spread over a greater number of hypothesized clusters. In linguistics and other fields, the DP has been widely used as a prior over lexica and other similar inventories \citep[e.g.,][]{Anderson90,KempPerforsTenenbaum07,Goldwater+06,Goldwater+09_word_seg,Teh06,Teh+06,Feldman+13,ODonnell15}.

\subsubsection{Likelihood}

We assume that each word is sampled from a probability distribution whose parameters are associated with the cluster to which the word belongs.\footnote{The parameters of each word cluster are optimized during the learning process together with the cluster assignments.} Then, the likelihood of a word partition is defined by the product of the probabilities of all words in the dataset given their cluster assignments defined by the partition.

Clusters with fewer words have less phonotactic variability and thus the probability distribution associated with the cluster is able to be more sharply peaked around a smaller set of phonotactic patterns assigning a higher probability to each word. For this reason, the likelihood favors finer-grained word partitions; in an extreme case, the likelihood is maximized when each individual word forms its own cluster, specialized to generate just that particular word. Thus, the likelihood and the prior have opposing preferences for word partitions, and learning amounts to balancing a tradeoff between these opposing pressures.

For the specific implementation of the likelihood, we utilize a trigram model of phoneme strings \citep[with hierarchical backed-off smoothing;][]{Goldwater+06,Teh06}. This model defines probabilities of phonemes conditioned on their two closest predecessors within words, and the overall word probability is the product of these phoneme probabilities. While trigram models may not capture all phonotactic patterns in the world's languages \citep{Hansson01,RoseWalker04,Heinz10},%
\footnote{%
    A higher likelihood might be achieved by employing an artificial neural network capable of modeling longer-range stochastic dependencies on preceding elements. Indeed, the current state-of-the-art model for time series data, Transformer, can be mathematically interpreted as an extended version of the $n$-gram model, allowing a substantially larger effective Markov order of $n-1$ \citep{Vaswani+17_AttentionIsAllYouNeed}. However, permitting long-distance dependencies among arbitrary segments is not appropriate for modeling phonotactics, since cross-linguistically, only phonetically related or similar sounds interact at a distance \citep[see General Discussion for more details]{HayesWilson08,MoritaODonnell_sublexical_phonotactics}. Furthermore, our simpler trigram model integrates more readily with the DP prior used in our Bayesian inference framework.}
they can effectively express local phonotactic dependencies among segments and account for a major portion of attested phonotactics \citep{Gafos99,NiChiosainPadgett01,HayesWilson08}. With this model, the likelihood of a word partition is simply the product of the probabilities of all words in the dataset given their cluster assignments defined by the partition.

\subsubsection{Bayesian Inference}

The optimal word clustering is formalized as the computation of the \emph{posterior} probability, proportional to the product of the prior and likelihood by Bayes' theorem. A similar approach to balancing between simplicity and fit to the data is inherent in various theories of inductive inference \citep[][]{RissanenRistad94,LiVitanyi08,grunwald.p:2007,shalev-shwartz.s:2014,vapnik.v:1998,clark.a:2011,jain.s:1999,bernardo.j:1994}.

A major challenge in this Bayesian inference is that the exact computation of the posterior probabilities is typically computationally intractable. Accordingly, we resort to variational approximation (specifically, the mean-field approximation) of the posterior to effectively handle the computational complexity and obtain practical solutions \citep[][]{Bishop06,BleiJordan06, WainwrightJordan08book, WainwrightJordan08,Wang+11,Blei+17}.

\subsection{Data} \label{sec:training-data}

The clustering method introduced above was applied to the (British) English portion of the CELEX lexical database \citep{CELEX}. We adopted the original phonetic transcription of the dataset (called DISC; see Appendix~\ref{sec:celex} for details) to represent the input.

Our model solely relies on the segmental information in words; accordingly, prosodic information---specifically, stress and syllable boundary markers---was removed from the transcriptions.%
\footnote{%
    There are several possibilities for extending our trigram model to incorporate prosodic information of words. See the Limitations section for details.}

Our data only distinguished lemmas, ignoring spelling and inflectional variations among words (such as singular-plural distinctions).%
\footnote{
    Some lemmas had more than one possible pronunciation; in such cases, we adopted the first (leftmost) entry.}
We further limited the data to lemmas with positive frequency in the Collins Birmingham University International Language Database \citep[COBUILD;][]{Sinclair87}. After filtering, there remained 38,731 words in the input data.

\section{Clustering Results} \label{sec:clustering-results}

\begin{table}
    \caption{Clustering results based on the maximum-a-posteriori (MAP) prediction of the model.}
    \label{tab:MAP-size}
    \centering
    \begin{tabular}{lrr}
        \toprule
        Cluster Name
        &
            \#Words
            &
                Proportion
                \\
        \midrule
        \textsc{Sublex}\textsubscript{$\approx$Germanic}
        &
            23,354
            &
                60.3\%
                \\
        \textsc{Sublex}\textsubscript{$\approx$Latinate}
        &
            15,174
            &
                39.2\%
                \\
        \textsc{Sublex}\textsubscript{$\approx$-ity}
        &
            203
            &
                0.5\%
                \\
    \bottomrule
    \end{tabular}
\end{table}

The unsupervised clustering revealed two primary sublexical clusters, labeled as \textsc{Sublex}\textsubscript{$\approx$Germanic} and \textsc{Sublex}\textsubscript{$\approx$Latinate} (Table~\ref{tab:MAP-size}). Additionally, a small cluster, labeled as \textsc{Sublex}\textsubscript{$\approx$-ity}, was identified, consisting of words that end with the suffix \textit{-ity}.%
\footnote{%
    All but one of the 203 words in \textsc{Sublex}\textsubscript{$\approx$-ity} were singular, with a single exception of \textit{susceptibilities}.}
The emergence of this minor cluster suggests that a significant proportion of English words are formed through the suffixation of \textit{-ity}, indicating the exceptional productivity of this suffix. Given that our model operates without awareness of morphological structures, there remains little else to discuss on \textsc{Sublex}\textsubscript{$\approx$-ity}. Therefore, the remainder of this paper is devoted to discussing linguistic properties of the other two major clusters, \textsc{Sublex}\textsubscript{$\approx$Germanic} and \textsc{Sublex}\textsubscript{$\approx$Latinate}.

\begin{figure}
    \centering
    \includegraphics[width=110mm]{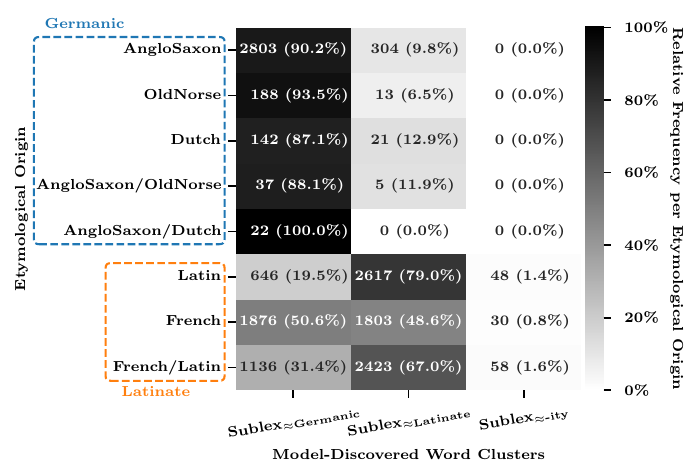}
    \caption{Alignment between the model-discovered clusters (columns, MAP classification) and the etymological origin according to Wikipedia (rows).}
    \label{fig:MAP-alignment}
\end{figure}

Figure~\ref{fig:MAP-alignment} illustrates the alignment between the discovered clusters (columns) and ground-truth etymological origins (rows). Due to the absence of etymological information in the CELEX dataset, we evaluated only a subset of the data (3,535 Germanic words and 10,637 Latinate words) whose etymological origin was identified in Wikipedia articles (see Appendix~\ref{sec:wiki} for details). Each cell of the heatmap is annotated with the word counts of the corresponding cluster-etymology intersection, followed by their relative frequency (in parentheses) per etymological origin (i.e., normalized over the columns, per row). The heatmap darkness also represents this relative frequency. The etymological origins are grouped into Germanic and Latinate by blue and orange dashed lines, respectively. The rows labeled with multiple origins (e.g., ``AngloSaxon/OldNorse'') represent the words duplicated in the Wikipedia lists of the corresponding origins.

Germanic words---of Anglo-Saxon, Old Norse, or Dutch origin---were closely aligned with the discovered \textsc{Sublex}{\textsubscript{$\approx$Germanic}} class. By contrast, the alignment between the discovered \textsc{Sublex}{\textsubscript{$\approx$Latinate}} class and words of etymologically Latinate origin showed less consistency; while Latin-derived words predominantly clustered into \textsc{Sublex}\textsubscript{$\approx$Latinate}, those of French origin were almost evenly split between the two clusters. However, this imperfect alignment of the model predictions with the ground-truth etymology is not necessarily a disappointing result; 
later in this work,
we will see that our model's ``misclassifications'' exhibit stronger correlations with the grammaticality of double-object constructions than the ground-truth etymology, thus providing a more effective account of the ``exceptions'' in the previous generalizations.

To quantitatively assess the significance of the alignment between our unsupervised clustering and the ground-truth etymology, we employed the V-measure metric \citep{RosebergHirschberg07}. The V-measure evaluates the similarity between predicted clustering and ground-truth classification based on two desiderata:
\begin{itemize}
    \item Homogeneity: each of the predicted clusters should contain only members of a single ground-truth class.
    \item Completeness: the members of each ground-truth class should be grouped into the same cluster.
\end{itemize}
Homogeneity and completeness are formally defined based on a normalized variant of the  conditional entropy, both falling on a scale of 0 (worst) to 1 (best). The V-measure score is their harmonic mean (analogous to the F1-score).

The V-measure score of our clustering result was $0.198$, significantly greater than the chance-level baseline drawn by random shuffling of the ground-truth word origins ($p < 10^{-5}$).%
\footnote{%
    The $p$-value was estimated using Monte Carlo: We sampled 100,000 random permutations of the ground-truth classifications, and the $p$-value was defined by the proportion of the random permutations whose V-measure score was greater than the model performance \citep{Ewens03_MC-p-val}. None of the 100,000 random permutations achieved a V-measure as great as the model, thus yielding $p < 10^{-5}$.}

\section{Phonotactic Characterization of the Discovered Word Clusters} \label{sec:phonotactic-representativeness}

In this section, we investigate the phonotactic properties of the model-detected clusters, \textsc{Sublex}\textsubscript{$\approx$Germanic} and \textsc{Sublex}\textsubscript{$\approx$Latinate}, examining if they are consistent with the observations made in the previous literature. 
We also conduct a data-driven exploration of the phonotactic features of these clusters, aiming to uncover previously unrecognized patterns. 

\subsection{Metric of Representativeness} \label{sec:methods--phonotactic-representativeness}

Following \cite{MoritaODonnell_sublexical_phonotactics}, our analysis of the phonotactic properties of the identified clusters is grounded in a metric of \emph{representativeness} of phonetic segments \citep{TenenbaumGriffiths01}. This metric assesses the relative likelihood that a sequence of phonetic segments comes from a particular cluster compared to the other(s). Essentially, representativeness is highest for patterns that are highly probable in the target cluster but improbable in the other(s). Consequently, it helps us identify (strings of) segments that provide informative cues for classification.

Formally, the representativeness $R(\mathbf{x}, k)$ of a string of phonetic segments $\mathbf{x} \allowbreak:=\allowbreak \left(x_1,\allowbreak\right. \dots,\allowbreak \left. x_m\right)$ with respect to the word cluster $k$ is defined by the logarithmic ratio of the posterior predictive probability of $\mathbf{x}$ appearing somewhere in a word belonging to the cluster $k$, to the average posterior predictive probability of $\mathbf{x}$ over all other clusters:
\begin{align}
    R(\mathbf{x}, k)
        :=
                \log
                    \frac{
                        p(
                            {\dots}\mathbf{x}{\dots}
                            \mid
                            k,
                            \mathbf{d}
                            )
                    }{
                        \sum_{k' \neq k}
                            p(
                                {\dots}\mathbf{x}{\dots}
                                \mid
                                k',
                                \mathbf{d}
                                )
                                p(
                                    k'
                                    \mid
                                    \mathbf{d},
                            k' \neq k
                                    )
                    }
                    \label{eq:representativeness}
\end{align}
where $\mathbf{d}$ denotes the training data of the clustering. For a detailed explanation of how the representativeness is computed, we refer interested readers to \citet{MoritaODonnell_sublexical_phonotactics}.

\subsection{Recovery of Previous Generalization Regarding Stress Patterns} \label{sec:topdown-results--phonotactic-representativeness}

We initiate our phonotactic analyses by showing that the model recovers generalizations proposed in the previous literature. Specifically, we discuss the prosodic characterization that Germanic verbs bear initial stress whereas Latinate verbs have unstressed initial syllable \citep{Grimshaw85,GrimshawPrince86}.

It is important to note that the training data for our model did not explicitly include prosodic information of words, such as syllables or stress patterns. Thus, the model cannot make direct predictions regarding the stress patterns of word-initial syllables. Nonetheless, the model can still make indirect predictions for the prosodic patterns, exploiting the fact that most of the English vowels are reduced to schwa [{\textschwa}] in unstressed positions. Specifically, we can evaluate which English vowels are most representative of \textsc{Sublex}\textsubscript{$\approx$Latinate} and \textsc{Sublex}\textsubscript{$\approx$Germanic} when they appear as the first vowel in a word. If schwa has a high degree of representativeness with respect to \textsc{Sublex}\textsubscript{$\approx$Latinate} in first position, it indicates that the unstressed word-initial syllable is representative of the word cluster.

We focus our analysis on the initial vowels of polysyllabic words,%
\footnote{%
    To compute the representativeness of a vowel $\textrm{V}_1$ appearing in the first syllable of a polysyllabic word, we replace $p(\dotsc\mathbf{x}\dotsc \mid k^{(\prime)}, \mathbf{d})$ in Eq.~\ref{eq:representativeness} with $p(\textrm{V}_1 \mid \textrm{C\textsubscript{1}\textsuperscript{*}\underline{\protect\phantom{V}}C\textsubscript{2}\textsuperscript{*}V\textsubscript{2}}, k^{(\prime)}, \mathbf{d})$: that is, the posterior predictive probability of $\textrm{V}_1$ conditioned on the context {C\textsubscript{1}\textsuperscript{*}\underline{\protect\phantom{V}}C\textsubscript{2}\textsuperscript{*}V\textsubscript{2}}, where {C\textsubscript{1}\textsuperscript{*}} and {C\textsubscript{2}\textsuperscript{*}} represent the existence of an arbitrary number of consonants (including ``no consonants'') in the positions and {V\textsubscript{2}} represents the existence of another vowel in the word.  In practice, we constrain {C\textsubscript{1}\textsuperscript{*}} up to three consonants and {C\textsubscript{2}\textsuperscript{*}} up to five consonants, based on the maximum length of the word-initial and internal consonant clusters in the CELEX data.}
simply because monosyllabic words always bear initial stress. Moreover, Germanic words are more likely to be monosyllabic than Latinate words \citep{Gropen+89}; thus, without the requirement of polysyllabicity, initial unstressed vowels can be representative of \textsc{Sublex}\textsubscript{$\approx$Latinate} merely due to the greater expected word length, derailing our primary interest in stress patterns.%
\footnote{%
    In our trigram model, the expected length of words is represented by the probability of a special symbol marking the word-final position.}

\begin{table}
    \caption{Representativeness of the first vowels in polysyllabic words (with zero to three initial consonants and zero to five internal consonants between the first and second vowels) with respect to \textsc{Sublex}\textsubscript{$\approx$Germanic} and \textsc{Sublex}\textsubscript{$\approx$Latinate}.}
    \label{tab:rep-init-Vs}
    \centering
    \begin{tabular}{lrr}
    \toprule
    Vowel & \textsc{Sublex}\textsubscript{$\approx$Germanic} & \textsc{Sublex}\textsubscript{$\approx$Latinate} \\
    \midrule
    {\textschwa} & -1.832312 & 1.720911 \\
    {\textsci} & -1.173901 & 1.136778 \\
    {\textupsilon}{\textschwa} & -0.819426 & 0.736090 \\
    {\textsci}{\textschwa} & -0.704950 & 0.691387 \\
    {\textepsilon} & -0.291054 & 0.292225 \\
    {\ae} & -0.162233 & 0.160080 \\
    {\textturnscripta} & -0.127456 & 0.135075 \\
    {\~{\textscripta}}{\textlengthmark} & -0.470889 & -0.043847 \\
    {\textrevepsilon}{\textlengthmark} & 0.063449 & -0.074478 \\
    {\~{\ae}} & -0.570518 & -0.184845 \\
    u{\textlengthmark} & 0.343961 & -0.341091 \\
    {\textopeno}{\textlengthmark} & 0.342331 & -0.342073 \\
    {\~{\ae}}{\textlengthmark} & -0.314203 & -0.396665 \\
    {\~{\textturnscripta}}{\textlengthmark} & -0.277864 & -0.426937 \\
    a{\textsci} & 0.529269 & -0.534647 \\
    {\textturnv} & 0.545830 & -0.537470 \\
    {\textschwa}{\textupsilon} & 0.584508 & -0.576595 \\
    {\textopeno}{\textsci} & 0.555049 & -0.587985 \\
    {\textscripta}{\textlengthmark} & 0.608208 & -0.597709 \\
    i{\textlengthmark} & 0.849564 & -0.840888 \\
    {\textupsilon} & 1.001237 & -0.998456 \\
    {\textepsilon}{\textschwa} & 1.141281 & -1.181892 \\
    e{\textsci} & 1.172851 & -1.183918 \\
    a{\textupsilon} & 1.697911 & -1.707058 \\
    \bottomrule
    \end{tabular}
\end{table}

Table~\ref{tab:rep-init-Vs} reports the representativeness scores of all English vowels with respect to \textsc{Sublex}\textsubscript{$\approx$Germanic} and \textsc{Sublex}\textsubscript{$\approx$Latinate} when occurring in the first syllable of polysyllabic words.%
\footnote{%
    Note that vowels can have negative representativeness for both \textsc{Sublex}\textsubscript{$\approx$Germanic} and \textsc{Sublex}\textsubscript{$\approx$Latinate} (e.g., {\~{\textscripta}}{\textlengthmark}) when they have positive representativeness with respect to \textsc{Sublex}\textsubscript{$\approx$-ity}.}
The phonetic transcription (of British English) was translated from DISC to IPA for readability.

Schwa [{\textschwa}] scored the lowest in \textsc{Sublex}\textsubscript{$\approx$Germanic} and the highest in \textsc{Sublex}\textsubscript{$\approx$Latinate}. This finding aligns with the previous generalization that initial stress (i.e., non-schwa initial vowels) is a hallmark of Germanic words.

\subsection{Data-Driven Investigation of Representative Phonotactic Patterns} \label{sec:bottomup-results--phonotactic-representativeness}

\begin{table*}
    \caption{Uni- to trigram substrings yielding the greatest representativeness. The non-IPA tokens, \texttt{END} and * (asterisk), represent the word-final position and linking \textit{r}, respectively.}
    \label{tab:top-5-representative-ngrams}
    \centering
    \begin{tabular}{lc
        p{1.5em}@{}p{1.5em}@{}p{1.75em}rl 
        p{1.5em}@{}p{1.5em}@{}p{1.75em}rl 
        }
        \toprule
            &
                \multirow{2}{*}{\rotatebox{90}{Rank}}&
                    \multicolumn{5}{l}{\textsc{Sublex}\textsubscript{$\approx$Germanic}}&
                    \multicolumn{5}{l}{\textsc{Sublex}\textsubscript{$\approx$Latinate}}\\
            &
                &
                    \multicolumn{3}{c}{Substring}&
                        Rep.\,Score&
                        Examples&
                    \multicolumn{3}{c}{Substring}&
                        Rep.\,Score&
                        Examples\\
        \midrule
            \multirow{5}{*}{\rotatebox{90}{Unigram}}&
                1&
                    \textipa{D}&&&
                        1.5379&
                            \textit{ba\underline{th}e}, \textit{mo\underline{th}er}&
                    \textsyllabic{n}&&&
                        1.9389&
                        \textit{esse\underline{n}ce}, \textit{natio\underline{n}}\\
            &
                2&
                    a{\textupsilon}&&&
                        1.4040&
                        \textit{ab\underline{ou}t}, \textit{l\underline{ou}d}&
                    {\textupsilon}{\textschwa}&&&
                        1.4407&
                        \textit{c\underline{u}re}, \textit{d\underline{u}ration}\\
            &
                3&
                    w&&&
                        1.3100&
                        \textit{\underline{w}ork}, \textit{\underline{w}ound}&
                    j&&&
                        1.2592&
                        \textit{acc\underline{u}se}, \textit{\underline{u}se}\\
            &
                4&
                    \textipa{N}&&&
                        1.2497&
                        \textit{bli\underline{n}k}, \textit{swi\underline{ng}}&
                    {\textsci}{\textschwa}&&&
                        0.9589&
                        \textit{r\underline{ea}r}, \textit{s\underline{e}r\underline{ia}l},\\
            &
                5&
                    h&&&
                        1.2479&
                        \textit{\underline{h}and}, \textit{\underline{h}ole}&
                    v&&&
                        0.8201&
                        \textit{sur\underline{v}i\underline{v}e}, \textit{vacation}\\
        \midrule
            \multirow{5}{*}{\rotatebox{90}{Bigram}}&
                1&
                    h & u{\textlengthmark}&&
                        4.7799&
                        \textit{\underline{hoo}p}, \textit{\underline{who}}&
                    {\textesh} & \textsyllabic{n}&&
                        5.5336&
                        \textit\textit{o\underline{cean}}, {suffi\underline{cien}t}\\
            &
                2&
                    h &{\textupsilon}&&
                        4.7651&
                        \textit{\underline{hoo}k}, \textit{likeli\underline{hoo}d}&
                    {\textteshlig} & {\textupsilon}{\textschwa}&&
                        4.4053&
                        \textit{ac\underline{tua}l}, \textit{vir\underline{tuou}s}\\
            &
                3&
                    f &{\textupsilon}&&
                        4.3186&
                        \textit{care\underline{fu}l}, \textit{\underline{foo}t}&
                    e{\textsci} & {\textesh}&&
                        4.3633&
                        \textit{f\underline{ac}ial}, \textit{r\underline{at}io}\\
            &
                4&
                    {\textepsilon}{\textschwa} & *&&
                        4.1726&
                        \textit{h\underline{air}}, \textit{th\underline{ere}}&
                    j & {\textupsilon}&&
                        3.8582&
                        \textit{occ\underline{u}py}, \textit{pop\underline{u}lar}\\
            &
                5&
                    r &a{\textupsilon}&&
                        4.0835&
                        \textit{b\underline{row}n}, \textit{g\underline{rou}nd}&
                    {\textyogh} & \textsyllabic{n}&&
                        3.7467&
                        \textit{deci\underline{sion}}, \textit{vi\underline{sion}}\\
        \midrule
            \multirow{5}{*}{\rotatebox{90}{Trigram}}&
                1&
                    \textipa{N} & {\textscriptg} & \textsyllabic{l}&
                        9.0861&
                        \textit{mi\underline{ngle}}, \textit{wra\underline{ngle}}&
                    e{\textsci} & {\textesh} & \textsyllabic{n}&
                        8.8990&
                        \textit{educ\underline{ation}}, \textit{p\underline{atien}t}\\
            &
                2&
                    h & a{\textupsilon} & s&
                        8.1806&
                        \textit{\underline{house}},
                        \textit{ware\underline{house}}&
                    j & {\textupsilon} & r&
                        8.6861&
                        \textit{acc\underline{ur}ate}, \textit{merc\underline{ur}y}\\
            &
                3&
                    i{\textlengthmark} & p & \texttt{END}&
                        7.8941&
                        \textit{d\underline{eep}}, \textit{sl\underline{eep}}&
                    p & {\textteshlig} & {\textupsilon}{\textschwa}&
                        7.8682&
                        \textit{conce\underline{ptua}l}, \textit{sum\underline{ptuou}s}\\
            &
                4&
                    {\textturnv} & {\textesh} & \texttt{END}&
                        7.8941&
                        \textit{fl\underline{ush}}, \textit{r\underline{ush}}&
                    k & {\textesh} & \textsyllabic{n}&
                        7.5597&
                        \textit{a\underline{ction}}, \textit{se\underline{ction}}\\
            &
                5&
                    k & {\textupsilon} & k&
                        7.3461&
                        \textit{\underline{cook}}, \textit{\underline{cook}ie}&
                    {\textesh} & \textsyllabic{n} & {\textschwa}&
                        7.5533&
                        \textit{sta\underline{tiona}ry}, \textit{na\underline{tiona}lism}\\
        \bottomrule
    \end{tabular}
\end{table*}

In addition to recovering the previous generalization of Germanic and Latinate phonology, our model can also be used for a data-driven investigation to identify class-specific phonotactic patterns. Specifically, ranking phonotactic patterns by their representativeness with respect to each cluster can unveil previously unnoticed characteristic patterns, suggesting new hypotheses for future experimental studies.

It is important to note that the representativeness-based analysis does not eliminate ungrammatical phonotactic patterns that never appear in English words. This is because our trigram phonotactic model is smoothed and assigns non-zero probabilities to unobserved patterns; consequently, a string of segments that is extremely improbable across all clusters can still be representative of one cluster if its probability in that cluster is relatively greater than in the others. Given the challenge of interpreting such ungrammatical (and rare) patterns, we limit our ranking to substrings with a minimum frequency of ten \citep[cf.][]{MoritaODonnell_sublexical_phonotactics}.

Table~\ref{tab:top-5-representative-ngrams} presents the top-five uni- to trigram substrings ranked by their representativeness.
\footnote{%
    The comprehensive list of all uni- to trigram substrings is also available as supplementary material.}
These substrings are largely consistent with our intuition. For instance, many of the high-ranking bigrams and trigrams for \textsc{Sublex}\textsubscript{$\approx$Latinate} correspond to the Latinate suffix \textit{-(at)ion}, as exemplified by [{\textyogh} \textsyllabic{n}] and [e{\textsci} {\textesh} \textsyllabic{n}].

Similarly, it is reasonable that the word final [i{\textlengthmark} p] (represented as a trigram [i{\textlengthmark} p \texttt{END}] in our model) is characteristic of \textsc{Sublex}\textsubscript{$\approx$Germanic}, given that most words exhibiting this phonotactic pattern are Germanic, such as \textit{cr\underline{eep}}, \textit{d\underline{eep}}, \textit{h\underline{eap}}, \textit{l\underline{eap}}, \textit{sl\underline{eep}}, \textit{sh\underline{eep}}, \textit{st\underline{eep}}, and \textit{sw\underline{eep}} \citep{NewOxfordAmericanDict3ed}.%
\footnote{%
    The only possible counterexample to this generalization is \textit{ch\underline{eap}}, which was built based on the Latin word \textit{caupo} `small trader, innkeeper'. This Latin word, however, was adopted in early Proto-Germanic \citep{ConciseOxfordDictionaryOfEnglishEtymology}, and thus is considered more adapted to the native phonotactics.}
Likewise, the most representative bigram [h u\textlengthmark] appears exclusively in Germanic words like \textit{\underline{hoo}p} and \textit{\underline{who}(m)}. Despite their plausibility, however, these phonetic characterizations of Germanic words had never been documented previously to our knowledge, indicating the effectiveness of data-driven investigation for identifying novel patterns.

\section{Word-Internal Consistency of Morpheme Etymology} \label{sec:base-suffix}
In the previous section, we phonotactically characterized the word clusters identified by our model. Conversely, our model infers these clusters based on the word-internal correlations among such phonotactic patterns; frequent cooccurrences of different substrings within words are better explained by a mixture of multiple distributions---each fitting to specific cooccurring patterns---rather than by a single trigram distribution, which assumes i.i.d. sampling (or mere coincidence) of the cooccurring substrings. Consequently, the presence of long words is essential for our model to observe sufficient cooccurrences.

Long words are typically formed through the affixation of morphemes. Therefore, successful learning of word clusters presupposes the word-internal consistency of phonotactic distributions across morphemes. In other words, our model exploits the etymological consistency across morphemes \citep[e.g., Latinate suffixes attach to Latinate bases;][]{Anshen+86,Fabb88,ODonnell15}. In this section, we demonstrate that this word-internal etymological consistency is indeed reflected in our model, by showing that it classifies different base morphemes of a common affix into the same cluster.

\begin{figure}
    \centering
    \includegraphics[width=110mm]{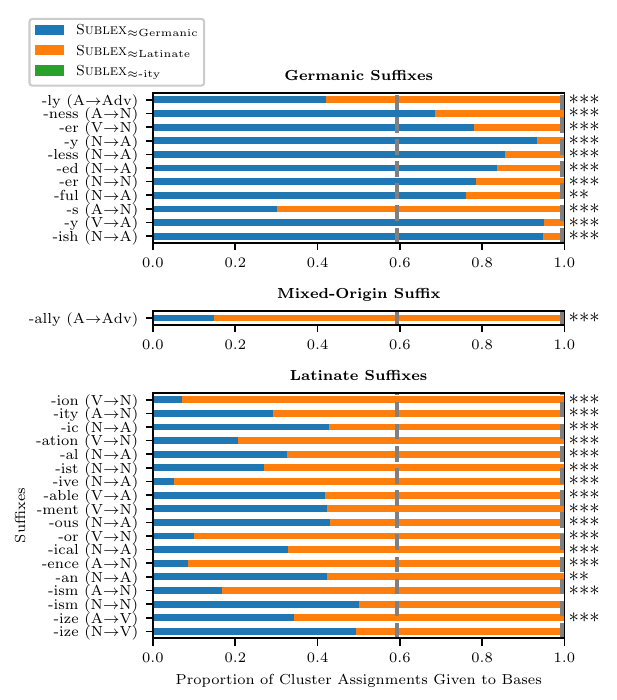}
    \caption{The proportion of the MAP cluster assignments given to the bases of the top thirty type-frequent suffixes.}
    \label{fig:suffixes}
\end{figure}

Figure~\ref{fig:suffixes} illustrates the proportion of cluster assignments given to the bases of the thirty most productive suffixes (e.g., the bases of \textit{-ion} include \textit{execute}, \textit{suspend}, \textit{profess}, etc.), ranked by the number of words derived through the suffixation (i.e., type frequency), as documented in the CELEX dataset. The vertical dashed lines indicate the proportion of the cluster assignments expected from the overall ratio (i.e., under the null hypothesis; \textsc{Sublex}\textsubscript{$\approx$Germanic}: 59.33\%, \textsc{Sublex}\textsubscript{$\approx$Latinate}: 40.13\%, \textsc{Sublex}\textsubscript{$\approx$-ty}: 0.53\%). The asterisks on the right to the bars represent the statistical significance of the deviation from the null hypothesis according to the multinomial test (\texttt{*}: $p<0.05$, \texttt{**}: $p<0.01$, \texttt{***}: $p<0.001$). The conversion of the grammatical category by the suffixation is annotated in the parentheses (e.g., ``A $\to$ Adv'' represents derivation of adjectives to adverbs). The suffix \textit{-ally}, identified as a single suffix in the CELEX, was parsed as a concatenation of Latinate (\textit{-al}) and Germanic (\textit{-ly}) suffixes, and thus labeled as ``Mixed-Origin''.

Eleven of these suffixes were of Germanic origin, and eighteen were of Latinate origin \citep{NewOxfordAmericanDict3ed}; the remaining suffix, \textit{-ally}, was analyzed as the concatenation of the Latinate suffix \textit{-al} and the Germanic suffix \textit{-ly}---although the CELEX treats it as a single morpheme---and thus, it was categorized separately as a ``mixed-origin'' suffix. The suffixes and their corresponding bases were identified according to the morphological structures provided in the CELEX, with the clustering based on the freestanding forms of the bases, which were also identified in the CELEX (otherwise excluded from the analysis).

The vertical dashed lines in the figure indicate the overall proportion of cluster assignments over the entire CELEX dataset (\textsc{Sublex}\textsubscript{$\approx$Germanic}: 59.33\%, \textsc{Sublex}\textsubscript{$\approx$Latinate}: 40.13\%, \textsc{Sublex}\textsubscript{$\approx$-ty}: 0.53\%). Using this overall ratio as the null-hypothetical parameters of the multinomial test,%
\footnote{
        We used the R function \texttt{multinomial.test} in the \texttt{EMT} package and executed the exact test.}
we assessed the statistical significance of the suffix-dependent tendencies of base clustering towards \textsc{Sublex}\textsubscript{$\approx$Germanic} or \textsc{Sublex}\textsubscript{$\approx$Latinate}.

Overall, our model systematically classified the bases of Germanic and Latinate suffixes into \textsc{Sublex}\textsubscript{$\approx$Germanic} and \textsc{Sublex}\textsubscript{$\approx$Latinate}, respectively, thereby recovering the generalizations documented in the previous literature with statistical significance \citep{Anshen+86,Fabb88,ODonnell15}. Exceptions were observed for the bases of the Germanic suffixes \textit{-ly} (e.g., \textit{slow}$\to$\textit{slow\underline{ly}}) and \textit{-s}%
\footnote{
    Although the CELEX database treats this suffix, \textit{-s}, as an adjective-to-noun derivational morpheme, it might be more appropriate to analyze it as the inflectional plural marker.}
(e.g., \textit{acoustic}$\to$\textit{acoustic\underline{s}}, \textit{odd}$\to$\textit{odd\underline{s}}),
which tended to be classified into \textsc{Sublex}\textsubscript{$\approx$Latinate}. In addition, two Latinate suffixes, \textit{-ism} (for noun-to-noun derivation) and \textit{-ize} (for noun-to-verb derivation), did not exhibit a statistically significant deviation from the null-hypothetical clustering ratio.

\section{Syntactic Predictions from the Clustering Results} \label{sec:DOC}
Finally, we assess our model's capability to predict the syntactic grammaticality of double-object constructions (hereinafter abbreviated as DOC) of dative verbs. It should be noted that DOC can be ungrammatical for various reasons; for example, DOC is not permitted with verbs that have certain types of semantics, such as communication of propositions and propositional attitudes (e.g., \textit{announce}, \textit{report}) and manner of speaking \citep[e.g., \textit{scream}, \textit{whisper};][]{Levin93}. The influence of verb semantics is further supported by an experimental study by \citet{Gropen+89}, which showed that nonce verbs denoting transfer of possession were judged more acceptable for DOC than those with other meanings. Beyond the lexical properties of individual verbs, DOC acceptability is also influenced by contextual and discourse factors, including the animacy, novelty, and length of the dative and accusative arguments \citep{Collins95,Bresnan07,Bresnan+07,Sinclair+22,Rathi+24}.

Again, our model relies purely on phonotactic information and cannot make any syntactic predictions per se. However, our model can indirectly infer the grammaticality by clustering dative verbs based on phonotactics and substituting the etymology-based generalizations with these model-detected clusters \citep[cf.][]{Gropen+89}.

Specifically, we evaluate the alignment of our model's clustering with the following distinction:
\begin{itemize}
    \item {\supcheckMark}DOC verbs:\\*
            Dative verbs that permit DOC (and the prepositional construction).
    \item *DOC-\textsc{Lat} verbs:\\*
            Dative verbs that are said to disallow DOC \emph{solely due to their Latinateness} \citep[i.e., no other factors can distinguish them from {\supcheckMark}DOC verbs;][]{Levin93}.
\end{itemize}
The test data for this assessment were derived from \citep{Levin93}.%
\footnote{
    {\supcheckMark}DOC verbs are listed in \textsection2.1, Ex.~115 of \citep{Levin93}; and *DOC-\textsc{Lat} verbs are found in \textsection2.1, Ex.~118a.}%
\footnote{
    \citeauthor{Levin93}'s~\citeyear{Levin93} list of *DOC-\textsc{Lat} verbs includes \textit{broadcast}, but its components, \textit{broad} and \textit{cast}, are in fact both of Germanic origin \citep{NewOxfordAmericanDict3ed}. Accordingly, we excluded it from our main analysis. As a side note, this example was also classified into \textsc{Sublex}\textsubscript{$\approx$Germanic}, so it did not contribute to adjudication between our model and the true etymology-based account (as both failed to predict the ungrammaticality of DOC).}
A critical aspect of these data is that not all of {\supcheckMark}DOC verbs are etymologically Germanic in fact; \emph{there are exceptional Latinate verbs that allow DOC}.%
\footnote{
    The etymological origin of the {\supcheckMark}DOC verbs were identified by referring to Wikipedia articles (see Appendix~\ref{sec:wiki} for details) and the New Oxford American Dictionary \citep{NewOxfordAmericanDict3ed}. Consequently, we excluded etymologically ambiguous words, listed as ``Latinates of Germanic origin'' in Wikipedia, from the analysis. We also excluded \textit{netmail} and \textit{telex} because they are compounds/blends of Germanic and Latinate morphemes.}
This empirical fact gives room for our model---by predicting these exceptions---to outperform the etymology-based generalization in the previous literature.

\begin{figure}
    \centering
    \includegraphics[width=110mm]{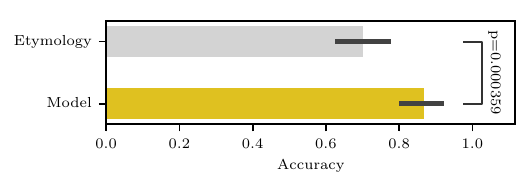}
    \caption{Accuracy scores of DOC grammatical prediction by the phonotactics-based clustering and the ground-truth etymology.}
    \label{fig:DOC-accuracy}
\end{figure}

Figure~\ref{fig:DOC-accuracy} reports the accuracy of distinguishing {\supcheckMark}DOC from *DOC-\textsc{Lat} verbs based on our clustering results and the true etymological classifications. Error bars indicate 95\% confidence intervals, estimated from 1000 bootstrap samples. The statistical significance of the difference between the model- and etymology-based predictions was assessed using the exact McNemar's test.

Remarkably, the model predictions (0.8681) surpassed those based on the true etymology (0.7014). As already noted above, this advantage is due to the fact that some of the Latinate verbs exceptionally permit DOC (termed {\supcheckMark}DOC-\textsc{Lat} hereinafter to distinguish them from the non-Latinate verbs, termed {\supcheckMark}DOC-$\overline{\textsc{Lat}}$) and our model ``correctly misclassified'' these exceptions to \textsc{Sublex}\textsubscript{$\approx$Germanic} based on their phonotactic patterns (Figure~\ref{fig:DOC_summary}; see Appendix~\ref{sec:DOC_full} for the clustering results of individual verbs). This outcome suggests that the grammaticality of DOC is more accurately generalized by phonotactic patterns rather than by the etymological origins. Indeed, this finding aligns with the experimental study by \citet{Gropen+89}, who showed the productivity of DOC utilizing monosyllabic and polysyllabic nonce words that characterized Germanic and Latinate verbs, respectively.

\begin{figure}
    \centering
    \includegraphics[width=110mm]{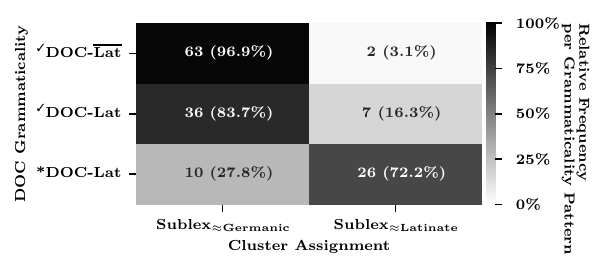}
    \caption{Alignment between the model-discovered clusters (columns, MAP classification) and the DOC grammaticality patterns (rows).
    Each cell of the heatmap is annotated with the word counts of the corresponding cluster-grammaticality intersection, followed by their relative frequency (in parentheses) per grammaticality pattern (i.e., normalized over the columns, per row). The heatmap darkness also represents this relative frequency.
    }
    \label{fig:DOC_summary}
\end{figure}

\section{General Discussion} \label{sec:discussion}
\subsection{Findings and Implications}

In this study, we demonstrated that the Germanic-Latinate distinction within the English lexicon is learnable from phonotactic information alone, without supervision. Specifically, we showed that the model-discovered clusters:
\begin{itemize}
    \item Broadly aligned with the ground-truth etymological classification,
    \item Recovered known phonotactic generalizations (stress patterns),
    \item Revealed hitherto unnoticed phonotactic properties (e.g., [i{\textlengthmark}p] and [hu{\textlengthmark}] as strong cues to membership in the Germanic-like cluster),
    \item Captured the etymological consistency of morphemes within words, and
    \item Predicted the grammaticality of DOC more accurately than the ground-truth etymology.
\end{itemize}
By empirically demonstrating the learnability of Latinate- and Germanic-like word clusters, the present study supports the psychological reality of linguistic theories that posit an etymological distinction in the mental grammar of native speakers \citep[e.g.,][]{ChomskyHalle68,ItoMester95JPhon}. Even though language learners do not have access to information about words' historical origins, they can acquire word clusters that approximate this distinction solely from phonotactic data. Moreover, the clusters identified by our model can capture linguistically meaningful generalizations more effectively, as evidenced by their superior performance in predicting DOC grammaticality.

The model predictions also offer opportunities for further experimental investigation into class-dependent linguistic properties. For example, researchers could test the psychological reality of the uni- to trigrams that our model identified as representative of the Latinate- and Germanic-like word clusters, employing experimental paradigms similar to those used in the previous studies \citep[e.g.,][]{MoretonAmano99}. Similarly, the clustering results could be leveraged to experimentally examine predictions about DOC grammaticality derived from our model \citep[cf.][]{Gropen+89}.

Finally, our findings support the cross-linguistic applicability of the proposed learning framework. \citet{MoritaODonnell_sublexical_phonotactics} demonstrated that the etymological word classes in Japanese can also be learned from phonotactic information using the same model, despite prior claims that such learning depends on orthographic cues unique to Japanese \citep[i.e., distinct writing systems are used for native vs. loanwords;][]{GelbartKawahara07}. Because the phonotactics-based approach requires only phonetic transcriptions, it can be applied to any spoken language, 
paving the way for future cross-linguistic tests of its universality.

Beyond cognitive-scientific inquiry, phonotactics-based clustering may also serve as a useful tool in historical linguistics, providing a starting point for identifying word etymology prior to more detailed analyses. It should be noted, however, that the method is entirely unsupervised; that is, alignments between the resulting clusters and true etymological classes are not guided by any supervision and therefore not guaranteed, as partially demonstrated by our DOC analysis.
Conversely, the possibility of such misalignments raises an intriguing question for future research: are there languages in which unsupervised clustering would reveal other kinds of word classes instead of etymological ones? For example, the grammatical gender of French nouns is largely predictable from their phonotactic patterns \citep[at least for a supervised classifier;][]{Lyster06}, and thus an unsupervised learner might discover gender-like clusters rather than etymology-like ones in such languages. We therefore expect the model to recover whichever lexical distinction is most strongly cued by phonotactics in a given language, rather than privileging etymology per se. Relatedly, our proposed model cannot represent multiple, partially overlapping lexical distinctions (e.g., etymology- and gender-based distinctions simultaneously). A possible extension to address this limitation would be to replace the DP with an Indian buffet process \citep[][]{GriffithsGhahramani06,GriffithsGhahramani11}, which allows words to have probabilistic membership in multiple latent classes.

\subsection{Alternative Approaches to Sublexicon Learning}

This study adopted a Bayesian framework to demonstrate the learnability of sublexical word clusters in the absence of supervision \citep{MoritaODonnell_sublexical_phonotactics}.
This section reviews alternative approaches to related problems explored in previous work.

\citet{ItoMester99} argued that the linguistic input available at an early stage of language acquisition may be restricted to words belonging to a specific sublexical class---particularly the native sublexicon---thereby leading learners to acquire phonotactic knowledge specialized for that class. Subsequently, learners may come to recognize the existence of additional sublexica when they encounter exceptions to this initially acquired phonotactic grammar. The plausibility of this staged-input hypothesis remains controversial. On the one hand, \citet{Gropen+89} observed that Latinate dative verbs were absent from spontaneous utterances of English-speaking children and were also rare in child-directed speech (both in either DOC and prepositional constructions). On the other hand, \citet{Ota99,Ota03,Ota04} reported that more than 50\% of the words produced by two-year-old Japanese children exhibited phonotactic patterns characteristic of non-native sublexica.

\citet{Pater05} proposed a learning algorithm that identifies words exhibiting exceptional phonotactic patterns and classifies them into a separate sublexicon from other, non-exceptional words. \citeauthor{Pater05} tested this algorithm only on idealized toy data---consisting of nine hypothetical words---and it has not been evaluated on an empirical dataset
\citep[see][for a more detailed discussion]{MoritaODonnell_sublexical_phonotactics}.

\citet{BeckerGouskova16} offered a probabilistic account of sublexicon-specific morpho-phonological patterns in Russian. They posited a classifier that groups masculine nominative nouns into two classes based on their phonotactic structure and predicts whether vowel deletion applies in their genitive forms according to the classification. This approach aligns with our analysis of DOC-possibility, abstracting away from the specific probability distributions assumed. For training, however, \citeauthor{BeckerGouskova16}'s sublexical classifier requires either supervision or explicit cues indicating the desired classification \citep[cf.][]{Ito+01}. In their case study of Russian, the nominative and genitive forms of each noun are assumed to be pre-identified, allowing the learner to infer the existence of two sublexica from the applicability of vowel deletion in the genitives. By contrast, our Bayesian method requires only sequences of phonetic segments as input data and can therefore be applied to any spoken language.

\subsection{Limitations and Possible Extensions of the Learning Model}
\label{sec:limitations}

We adopted an $n$-gram model of phonotactics to evaluate the likelihood of sublexical clusters. While this model can exhaustively capture local phonotactic dependencies among $n$ adjacent segments---accounting for a substantial portion of attested patterns \citep{Gafos99,NiChiosainPadgett01,HayesWilson08}---some languages exhibit unbounded long-distance dependencies \citep[e.g., anteriority agreement between sibilants in Navajo;][]{SapirHoijer67}.

To accommodate such long-distance dependencies, the Bayesian framework for sublexicon learning can be extended by incorporating a more expressive phonotactic model. For instance, \citet{Futrell+phonotactics} extended the $n$-gram model to capture distant dependencies between segments that share phonological features. Similarly, our learning paradigm is compatible with Optimality-Theoretic models, provided that they define a probability distribution over word forms \citep{HayesWilson08,Daland+11}.%
\footnote{
    Unlike $n$-gram and other autoregressive models (e.g., recurrent neural networks and Transformers), Optimality-Theoretic models typically define a probability distribution with a limited number of parameters (i.e., constraints). Consequently, they may fail to capture some of the stochastic differences among word clusters discovered by our model.}

Long-distance dependencies can also be modeled using modern neural-network architectures \citep{Vaswani+17_AttentionIsAllYouNeed,GuDao24}. However, these models permit unrestricted interactions among arbitrary phonemes, which conflicts with cross-linguistic evidence that only phonetically related or similar segments interact at a distance \citep{HayesWilson08,MoritaODonnell_sublexical_phonotactics}. For this reason, feature-based models constitute a more linguistically plausible direction for extending the present framework.

Feature-based models may also identify sublexical differences in local dependencies more efficiently. While the $n$-gram model treats individual segments as arbitrary symbols, feature-based models exploit inherent phonetic similarities and differences among segments \citep[see][for an empirical comparison]{Futrell+phonotactics}.%
\footnote{
    Conversely, some researchers have argued that phonological features themselves emerge from statistical learning rather than being inherent to individual phonemes. \citet{Nelson22}, for example, proposed an algorithm that groups phonemes based on their cooccurrence statistics, whose results aligned with traditional phonological features. This approach resembles the $n$-gram model, though its learned representations do not necessarily correspond to interpretable phonological features.}

Another limitation of our study is its lack of prosodic representations \citep{KagerPater12}. Although our model indirectly captured the difference in stress patterns between Germanic and Latinate verbs by exploiting segmental alternations associated with stress (vowel reduction), this strategy would not generalize to other languages. The extended $n$-gram model proposed by \citet{Futrell+phonotactics} addresses this issue by encoding prosodic information as suprasegmental features of vowels. Alternatively, one might adopt a hierarchical generative model that represents latent prosodic structures \citep{Lee+15}.

It should be noted, however, that all of the potential extensions discussed above are computationally more demanding than the $n$-gram model. This issue is particularly acute in the context of word-cluster learning, where multiple phonotactic models must be instantiated, one for each cluster. Indeed, even our $n$-gram model required a dedicated approximation method for Bayesian inference \citep[variational inference;][]{Bishop06,BleiJordan06, WainwrightJordan08book, WainwrightJordan08,Wang+11,Blei+17} to make full-scale analysis of the entire lexicon computationally feasible,%
\footnote{
    Bayesian posterior distributions can also be estimated using Markov Chain Monte Carlo (MCMC) methods \citep{Hastings70,Gilks+96,Neal00}. However, these sampling-based approaches proved impractical for our clustering problem under realistic time constraints.}
Because this approximation method was specifically tailored to the $n$-gram phonotactic model, adapting it to other phonotactic models requires customized formulations of the approximate posterior, which is technically non-trivial and thus poses a major challenge for future investigations of more advanced phonotactic models.

Beyond the model design and data representation, the full-batch learning paradigm adopted in this study---assuming that the learner has simultaneous access to the entire English vocabulary---is cognitively unrealistic. In reality, children acquire their vocabulary incrementally, and it remains an open question how word clusters might be identified under such an online learning scenario. In particular, the rarity of words from certain sublexica in child-directed speech may hinder the statistical inference of the corresponding word clusters \citep[e.g., Latinate words in English;][]{Gropen+89}.

\section*{Acknowledgments}
\lccode`\0`\0
\lccode`\1`\1
\lccode`\2`\2
\lccode`\3`\3
\lccode`\4`\4
\lccode`\5`\5
\lccode`\6`\6
\lccode`\7`\7
\lccode`\8`\8
\lccode`\9`\9
\hyphenation{JP-24-H-0-0-7-7-4 JP-22-H-0-3-9-1-4 JP-24-K-1-5-0-8-7 JP-MJCR-25-U6 JP-MJCR-22-P5 JP-21-K-1-7-8-0-5 K35-XXVIII-620 Kaya-mori}
This study was supported by JST AIP Accelerated Program (JPMJCR25U6), ACT-X (JPMJAX21AN), and CREST (JPMJCR22P5);
JSPS Grant-in-Aid for Early-Career Scientists (JP21K17805) 
and for Scientific Research A (JP24H00774), B (JP22H03914), and C (JP24K15087);
and Kayamori Foundation of Informational Science Advancement (K35XXVIII620).
We also gratefully acknowledge the support of the Canada CIFAR AI Chairs Program and the Natural Sciences and Engineering Research Council of Canada (NSERC).

\section*{Author Contributions}
TM and TJO jointly contributed to the conceptualization and writing of this study.
TM implemented the learning simulation and performed the analyses.
TJO supervised this project.

\bibliography{sublex_references}

\begin{thebibliography}{}

\bibitem[Anderson, 1990]{Anderson90}
Anderson, J.~R. (1990).
\newblock {\em The adaptive character of thought}.
\newblock Studies in cognition. L. Erlbaum Associates, Hillsdale, NJ.

\bibitem[Anshen et~al., 1986]{Anshen+86}
Anshen, F., Aronoff, M., Byrd, R., and Klavans, J. (1986).
\newblock The role of etymology and word-length in {English} word formation.
\newblock M.S.

\bibitem[Antoniak, 1974]{Antoniak74}
Antoniak, C.~E. (1974).
\newblock Mixtures of {Dirichlet} processes with applications to {Bayesian} nonparametric problems.
\newblock {\em The Annals of Statistics}, 2(6):1152--1174.

\bibitem[Baayen et~al., 1995]{CELEX}
Baayen, H.~R., Piepenbrock, R., and Gulikers, L. (1995).
\newblock {\em The {CELEX} Lexical Database. Release 2}.
\newblock {L}inguistic {D}ata {C}onsortium, University of Pennsylvania, Philadelphia, Pennsylvania.

\bibitem[Becker and Gouskova, 2016]{BeckerGouskova16}
Becker, M. and Gouskova, M. (2016).
\newblock Source-oriented generalizations as grammar inference in russian vowel deletion.
\newblock {\em Linguistic Inquiry}, 47(3):391--425.

\bibitem[Bernardo and Smith, 1994]{bernardo.j:1994}
Bernardo, J.~M. and Smith, A. F.~M. (1994).
\newblock {\em Bayesian Theory}.
\newblock John Wiley and Sons, Chichester.

\bibitem[Bishop, 2006]{Bishop06}
Bishop, C.~M. (2006).
\newblock {\em Pattern recognition and machine learning}.
\newblock Information science and statistics. Springer, New York.

\bibitem[Blei and Jordan, 2006]{BleiJordan06}
Blei, D.~M. and Jordan, M.~I. (2006).
\newblock Variational inference for {Dirichlet} process mixtures.
\newblock {\em Bayesian Analysis}, 1(1):121--144.

\bibitem[Blei et~al., 2017]{Blei+17}
Blei, D.~M., Kucukelbir, A., and McAuliffe, J.~D. (2017).
\newblock Variational inference: A review for statisticians.
\newblock {\em Journal of the American Statistical Association}, 112(518):859--877.

\bibitem[Bobaljik, 1997]{Bobaljik97}
Bobaljik, J.~D. (1997).
\newblock Mostly predictable: Cyclicity and the distribution of schwa in {Itelmen}.
\newblock In {\em Proceedings of the Twenty-Sixth Western Conference on Linguistics ({WECOL})}, pages 14--28.

\bibitem[Bresnan, 2007]{Bresnan07}
Bresnan, J. (2007).
\newblock Is syntactic knowledge probabilistic? experiments with the {English} dative alternation.
\newblock In Featherston, S. and Sternefeld, W., editors, {\em Roots: Linguistics in Search of Its Evidential Base}, pages 77--96. Mouton de Gruyter, Berlin.

\bibitem[Bresnan et~al., 2007]{Bresnan+07}
Bresnan, J., Cueni, A., Nikitina, T., and Baayen, H. (2007).
\newblock Predicting the dative alternation.
\newblock In Bouma, G., Kr{\"a}mer, I., and Zwarts, J., editors, {\em Cognitive Foundations of Interpretation}, pages 69--94, Amsterdam. Edita-the Publishing House of the Royal.

\bibitem[Brown et~al., 2020]{Brown+20}
Brown, T.~B., Mann, B., Ryder, N., Subbiah, M., Kaplan, J., Dhariwal, P., Neelakantan, A., Shyam, P., Sastry, G., Askell, A., Agarwal, S., Herbert-Voss, A., Krueger, G., Henighan, T., Child, R., Ramesh, A., Ziegler, D.~M., Wu, J., Winter, C., Hesse, C., Chen, M., Sigler, E., Litwin, M., Gray, S., Chess, B., Clark, J., Berner, C., McCandlish, S., Radford, A., Sutskever, I., and Amodei, D. (2020).
\newblock Language models are few-shot learners.
\newblock In Larochelle, H., Ranzato, M., Hadsell, R., Balcan, M.~F., and Lin, H., editors, {\em Advances in Neural Information Processing Systems}, volume~33, pages 1877--1901. Curran Associates, Inc.

\bibitem[Chomsky and Halle, 1968]{ChomskyHalle68}
Chomsky, N. and Halle, M. (1968).
\newblock {\em The sound pattern of {English}}.
\newblock Harper {\&} Row, New York.

\bibitem[Chung, 1983]{Chung83}
Chung, S. (1983).
\newblock Transderivational relationships in {Chamorro} phonology.
\newblock {\em Language}, 59(1):35--66.

\bibitem[Clark and Lappin, 2011]{clark.a:2011}
Clark, A. and Lappin, S. (2011).
\newblock {\em Linguistic Nativism and the Poverty of the Stimulus}.
\newblock Wiley-Blackwell.

\bibitem[Collins, 1995]{Collins95}
Collins, P. (1995).
\newblock The indirect object construction in english: an informational approach.
\newblock {\em Linguistics}, 33(1):35--50.

\bibitem[Daland et~al., 2011]{Daland+11}
Daland, R., Hayes, B., White, J., Garellek, M., Davis, A., and Norrmann, I. (2011).
\newblock Explaining sonority projection effects.
\newblock {\em Phonology}, 28(2):197--234.

\bibitem[Dingemanse, 2012]{Dingemanse12}
Dingemanse, M. (2012).
\newblock Advances in the cross-linguistic study of ideophones.
\newblock {\em Language and Linguistics Compass}, 6(10):654--672.

\bibitem[Dunbar et~al., 2017]{Dunbar+17_ZeroSpeech2017}
Dunbar, E., Cao, X.~N., Benjumea, J., Karadayi, J., Bernard, M., Besacier, L., Anguera, X., and Dupoux, E. (2017).
\newblock The zero resource speech challenge 2017.
\newblock In {\em 2017 {IEEE} Automatic Speech Recognition and Understanding Workshop ({ASRU})}, pages 323--330.

\bibitem[Ewens, 2003]{Ewens03_MC-p-val}
Ewens, W.~J. (2003).
\newblock On estimating p values by the monte carlo method.
\newblock {\em American journal of human genetics}, 72(2):496--498.

\bibitem[Fabb, 1988]{Fabb88}
Fabb, N. (1988).
\newblock English suffixation is constrained only by selectional restrictions.
\newblock {\em Natural language and linguistic theory}, 6(4):527--539.

\bibitem[Feldman et~al., 2013]{Feldman+13}
Feldman, N.~H., Goldwater, S., Griffiths, T.~L., and Morgan, J.~L. (2013).
\newblock A role for the developing lexicon in phonetic category acquisition.
\newblock {\em Psychological Review}, 120(4):751--778.

\bibitem[Feldman et~al., 2009]{Feldman+09}
Feldman, N.~H., Griffiths, T.~L., and Morgan, J.~L. (2009).
\newblock The influence of categories on perception: Explaining the perceptual magnet effect as optimal statistical inference.
\newblock {\em Psychological review}, 116(4):752--782.

\bibitem[Ferguson, 1973]{Ferguson73}
Ferguson, T.~S. (1973).
\newblock A {Bayesian} analysis of some nonparametric problems.
\newblock {\em The Annals of Statistics}, 1(2):209--230.

\bibitem[Frellesvig, 2010]{Frellesvig10}
Frellesvig, B. (2010).
\newblock {\em A History of the Japanese Language}.
\newblock Cambridge University Press, Cambridge.

\bibitem[Fries and Pike, 1949]{FriesPike49}
Fries, C.~C. and Pike, K.~L. (1949).
\newblock Coexistent phonemic systems.
\newblock {\em Language}, 25:29--50.

\bibitem[Fukazawa, 1998]{Fukazawa98}
Fukazawa, H. (1998).
\newblock Multiple input-output faithfulness relations in {J}apanese.
\newblock Rutgers Optimality Archive ROA-260-0698.

\bibitem[Fukazawa et~al., 1998]{Fukazawa+98_lexicalstratification}
Fukazawa, H., Kitahara, M., and Ota, M. (1998).
\newblock Lexical stratification and ranking invariance in constraint-based grammars.
\newblock In {\em CLS 34: The Panels}, pages 47--62.

\bibitem[Futrell et~al., 2017]{Futrell+phonotactics}
Futrell, R., Albright, A., Graff, P., and O'Donnell, T.~J. (2017).
\newblock A generative model of phonotactics.
\newblock {\em Transactions of the Association for Computational Linguistics}, 5:73--86.

\bibitem[Gafos, 1999]{Gafos99}
Gafos, A.~I. (1999).
\newblock {\em The Articulatory Basis of Locality in Phonology}.
\newblock Outstanding Dissertations in Linguistics. Routledge.

\bibitem[Gelbart and Kawahara, 2007]{GelbartKawahara07}
Gelbart, B. and Kawahara, S. (2007).
\newblock Lexical cues to foreignness in {Japanese}.
\newblock In Miyamoto, Y. and Ochi, M., editors, {\em Formal Approaches to Japanese Linguistics (FALJ 4)}, volume~55 of {\em MIT Working Papers in Linguistics}, pages 49--60, Cambridge, MA. MIT Working Papers in Linguistics.

\bibitem[Gilks et~al., 1996]{Gilks+96}
Gilks, W.~R., Richardson, S., and Spiegelhalter, D.~J. (1996).
\newblock Introducing {Markov} chain {Monte} {Carlo}.
\newblock In Gilks, W.~R., Richardson, S., and Spiegelhalter, D.~J., editors, {\em {Markov} chain {Monte} {Carlo} in Practice}, chapter~1, pages 1--19. Chapman {\&} Hall, London.

\bibitem[Goldwater et~al., 2006]{Goldwater+06}
Goldwater, S., Griffiths, T.~L., and Johnson, M. (2006).
\newblock Interpolating between types and tokens by estimating power-law generators.
\newblock In Weiss, Y., Sch{\"o}lkopf, B., and Platt, J.~C., editors, {\em Advances in Neural Information Processing Systems 18}, pages 459--466, Cambridge, MA. MIT {P}ress.

\bibitem[Goldwater et~al., 2009]{Goldwater+09_word_seg}
Goldwater, S., L~Griffiths, T., and Johnson, M. (2009).
\newblock A {Bayesian} framework for word segmentation: Exploring the effects of context.
\newblock {\em Cognition}, 112:21--54.

\bibitem[Goodenough and Sugita, 1990]{GoodenoughSuigita80}
Goodenough, W.~H. and Sugita, H. (1980/1990).
\newblock {\em {Trukese}-{English} Dictionary}.
\newblock American Philosophical Society.

\bibitem[Griffiths and Ghahramani, 2006]{GriffithsGhahramani06}
Griffiths, T.~L. and Ghahramani, Z. (2006).
\newblock Infinite latent feature models and the indian buffet process.
\newblock In Weiss, Y., Sch\"{o}lkopf, B., and Platt, J.~C., editors, {\em Advances in Neural Information Processing Systems 18}, pages 475--482. MIT Press.

\bibitem[Griffiths and Ghahramani, 2011]{GriffithsGhahramani11}
Griffiths, T.~L. and Ghahramani, Z. (2011).
\newblock The indian buffet process: An introduction and review.
\newblock {\em J. Mach. Learn. Res.}, 12:1185--1224.

\bibitem[Grimshaw, 1985]{Grimshaw85}
Grimshaw, J. (1985).
\newblock Remarks on dative verbs and universal grammar.
\newblock Presented at the 10th Annual Boston University Conference on Language Development.

\bibitem[Grimshaw and Prince, 1986]{GrimshawPrince86}
Grimshaw, J. and Prince, A. (1986).
\newblock A prosodic account of the {\it to}-dative alternation.
\newblock M.S.

\bibitem[Gropen et~al., 1989]{Gropen+89}
Gropen, J., Pinker, S., Hollander, M., Goldberg, R., and Wilson, R. (1989).
\newblock The learnability and acquisition of the dative alternation in english.
\newblock {\em Language}, 65(2):203--257.

\bibitem[Gr\"{u}nwald, 2007]{grunwald.p:2007}
Gr\"{u}nwald, P.~D. (2007).
\newblock {\em The Minimum Description Length Principle}.
\newblock The MIT Press, Cambridge, MA.

\bibitem[Gu and Dao, 2024]{GuDao24}
Gu, A. and Dao, T. (2024).
\newblock Mamba: Linear-time sequence modeling with selective state spaces.
\newblock In {\em Proceedings of the First Conference on Language Modeling}.

\bibitem[Hansson, 2001]{Hansson01}
Hansson, G. (2001).
\newblock {\em Theoretical and Typological Issues in Consonant Harmony}.
\newblock PhD thesis, University of California, Berkeley.

\bibitem[Hastings, 1970]{Hastings70}
Hastings, W.~K. (1970).
\newblock {Monte Carlo} sampling methods using {Markov} chains and their applications.
\newblock {\em Biometrika}, 57(1):97--109.

\bibitem[Hayes and Wilson, 2008]{HayesWilson08}
Hayes, B. and Wilson, C. (2008).
\newblock A maximum entropy model of phonotactics and phonotactic learning.
\newblock {\em Linguistic Inquiry}, 39(3):379--440.

\bibitem[Heinz, 2010]{Heinz10}
Heinz, J. (2010).
\newblock Learning long-distance phonotactics.
\newblock {\em Linguistic Inquiry}, 41(4):623--661.

\bibitem[Hoad, 2003]{ConciseOxfordDictionaryOfEnglishEtymology}
Hoad, T.~F. (1986/2003).
\newblock {\em The Concise Oxford Dictionary of English Etymology}.
\newblock Oxford University Press, Oxford; New York.

\bibitem[Holden, 1976]{Holden76}
Holden, K. (1976).
\newblock Assimilation rates of borrowings and phonological productivity.
\newblock {\em Language}, 52(1):131--147.

\bibitem[Hsu and Jesney, 2017]{HsuJesney17}
Hsu, B. and Jesney, K. (2017).
\newblock Loanword adaptation in {Q}u{\'e}bec {F}rench: Evidence for weighted scalar constraints.
\newblock In {\em Proceedings of the 34th West Coast Conference on Formal Linguistics}.

\bibitem[Hsu and Jesney, 2018]{HsuJesney18}
Hsu, B. and Jesney, K. (2018).
\newblock Weighted scalar constraints capture the typology of loanword adaptation.
\newblock In {\em Proceedings of the Annual Meetings on Phonology}, volume~5. Linguistic Society of America.

\bibitem[Ito and Mester, 1995a]{ItoMester95lexicon}
Ito, J. and Mester, A. (1995a).
\newblock The core-periphery structure of the lexicon and constraints on reranking.
\newblock In Beckman, J., Urbanczyk, S., and Walsh, L., editors, {\em Papers in {O}ptimality {T}heory {I}{I}{I}: {U}niversity of {M}assachusetts Occasional Papers 32}, pages 181--210. GLSA Publications, Amherst, MA.

\bibitem[Ito and Mester, 1995b]{ItoMester95JPhon}
Ito, J. and Mester, A. (1995b).
\newblock {J}apanese phonology.
\newblock In Goldsmith, J., editor, {\em A Handbook of Phonological Theory}, pages 817--838. Blackwell, Oxford.

\bibitem[Ito and Mester, 1999]{ItoMester99}
Ito, J. and Mester, A. (1999).
\newblock The phonological lexicon.
\newblock In Tsujimura, N., editor, {\em The handbook of {J}apanese linguistics}, pages 62--100. Blackwell, Oxford.

\bibitem[Ito and Mester, 2003]{ItoMester03}
Ito, J. and Mester, A. (2003).
\newblock Lexical and postlexical phonology in {O}ptimality {T}heory: evidence from {J}apanese.
\newblock {\em Linguistische Berichte}, 11:183--207.

\bibitem[Ito and Mester, 2008]{ItoMester08}
Ito, J. and Mester, A. (2008).
\newblock Lexical classes in phonology.
\newblock In Miyagawa, S. and Saito, M., editors, {\em The {O}xford handbook of {J}apanese linguistics}, chapter~4, pages 84--106. Oxford University Press, Oxford.

\bibitem[Ito et~al., 2001]{Ito+01}
Ito, J., Mester, A., and Padgett, J. (2001).
\newblock Alternations and distributional patterns in {Japanese} phonology.
\newblock {\em Journal of the Phonetic Society of Japan}, 5(2):54--60.

\bibitem[Jain et~al., 1999]{jain.s:1999}
Jain, S., Osherson, D., Royer, J.~S., and Sharma, A. (1999).
\newblock {\em Systems that Learn}.
\newblock The MIT Press, Cambridge, MA.

\bibitem[Kager and Pater, 2012]{KagerPater12}
Kager, R. and Pater, J. (2012).
\newblock Phonotactics as phonology: knowledge of a complex restriction in {Dutch}.
\newblock {\em Phonology}, 29(1):81--111.

\bibitem[Kelly and Bock, 1988]{KellyBock88}
Kelly, M.~H. and Bock, J.~K. (1988).
\newblock Stress in time.
\newblock {\em Journal of Experimental Psychology: Human Perception and Performance}, 14(3):389--403.

\bibitem[Kemp et~al., 2007]{KempPerforsTenenbaum07}
Kemp, C., Perfors, A., and Tenenbaum, J. (2007).
\newblock Learning overhypotheses with hierarchical {Bayesian} models.
\newblock {\em Developmental Science}, 10(3):307--321.

\bibitem[Lee et~al., 2015]{Lee+15}
Lee, C.-y., O'Donnell, T.~J., and Glass, J. (2015).
\newblock Unsupervised lexicon discovery from acoustic input.
\newblock {\em Transactions of the Association for Computational Linguistics}, 3:389--403.

\bibitem[Lees, 1961]{Lees61}
Lees, R.~B. (1961).
\newblock {\em The phonology of modern standard {Turkish}}.
\newblock Uralic and Altaic series. Indiana University publications.

\bibitem[Levin, 1993]{Levin93}
Levin, B. (1993).
\newblock {\em English Verb Classes and Alternations: A Preliminary Investigation}.
\newblock University of Chicago Press, Chicago; London.

\bibitem[Li and Vit{\'a}nyi, 2008]{LiVitanyi08}
Li, M. and Vit{\'a}nyi, P.~M. (2008).
\newblock {\em An Introduction to {Kolmogorov} Complexity and Its Applications}.
\newblock Texts in Computer Science. Springer, New York, NY, USA.

\bibitem[Lyster, 2006]{Lyster06}
Lyster, R. (2006).
\newblock Predictability in french gender attribution: A corpus analysis.
\newblock {\em Journal of French Language Studies}, 16(1):69--92.

\bibitem[Meyers, 1997]{Meyers97}
Meyers, S. (1997).
\newblock {OCP} effects in {Optimality} {Theory}.
\newblock {\em Natural Language \& Linguistic Theory}, 15(4):847--892.

\bibitem[Mikolov et~al., 2013a]{Mikolov+13_word2vec}
Mikolov, T., Chen, K., Corrado, G.~S., and Dean, J. (2013a).
\newblock Efficient estimation of word representations in vector space.
\newblock In Bengio, Y. and LeCun, Y., editors, {\em Proceedings of the 1st International Conference on Learning Representations ({ICLR})}, Scottsdale, Arizona.

\bibitem[Mikolov et~al., 2013b]{Mikolov+13_NIPS}
Mikolov, T., Sutskever, I., Chen, K., Corrado, G.~S., and Dean, J. (2013b).
\newblock Distributed representations of words and phrases and their compositionality.
\newblock In Burges, C. J.~C., Bottou, L., Welling, M., Ghahramani, Z., and Weinberger, K.~Q., editors, {\em Advances in Neural Information Processing Systems}, volume~26. Curran Associates, Inc.

\bibitem[Moreton and Amano, 1999]{MoretonAmano99}
Moreton, E. and Amano, S. (1999).
\newblock Phonotactics in the perception of {Japanese} vowel length: Evidence for long-distance dependencies.
\newblock In {\em Proceedings of the 6th European Conference on Speech Communication and Technology}, pages 2679--2682.

\bibitem[Morita, 2018]{Morita_thesis}
Morita, T. (2018).
\newblock {\em Unsupervised Learning of Lexical Subclasses from Phonotactics}.
\newblock PhD thesis, Massachusetts Institute of Technology, Cambridge, MA.

\bibitem[Morita and O'Donnell, 2022]{MoritaODonnell_sublexical_phonotactics}
Morita, T. and O'Donnell, T.~J. (2022).
\newblock Statistical evidence for learnable lexical subclasses in {Japanese}.
\newblock {\em Linguistic Inquiry}, 53(1):87--120.

\bibitem[Neal, 2000]{Neal00}
Neal, R.~M. (2000).
\newblock {Markov} chain sampling methods for {Dirichlet} process mixture models.
\newblock {\em Journal of Computational and Graphical Statistics}, 9(2):249--265.

\bibitem[Nelson, 2022]{Nelson22}
Nelson, M.~A. (2022).
\newblock {\em Phonotactic Learning with Distributional Representations}.
\newblock PhD thesis, University of Massachusetts Amherst.

\bibitem[N{\'\i}~Chios{\'a}in and Padgett, 2001]{NiChiosainPadgett01}
N{\'\i}~Chios{\'a}in, M. and Padgett, J. (2001).
\newblock Markedness, segment realization, and locality in spreading.
\newblock In Lombardi, L., editor, {\em Constraints and Representations: Segmental Phonology in Optimality Theory}, pages 118--156. Cambridge University Press, Cambridge.

\bibitem[O'Donnell, 2015]{ODonnell15}
O'Donnell, T.~J. (2015).
\newblock {\em Productivity and reuse in language : a theory of linguistic computation and storage.}
\newblock MIT {P}ress, Cambridge, MA; London, England.

\bibitem[Ota, 1999]{Ota99}
Ota, M. (1999).
\newblock {\em Phonological Theory and the Acquisition of Prosodic Structure: Evidence from child {Japanese}}.
\newblock PhD thesis, Georgetown University.

\bibitem[Ota, 2003]{Ota03}
Ota, M. (2003).
\newblock {\em The Development of Prosodic Structure in Early Words: Continuity, Divergence and Change (Language Acquisition and Language Disorders)}.
\newblock John Benjamins.

\bibitem[Ota, 2004]{Ota04}
Ota, M. (2004).
\newblock The learnability of the stratified phonological lexicon.
\newblock {\em Journal of {J}apanese Linguistics}, 20(4):19--40.

\bibitem[Ouyang et~al., 2022]{Ouyang+22}
Ouyang, L., Wu, J., Jiang, X., Almeida, D., Wainwright, C.~L., Mishkin, P., Zhang, C., Agarwal, S., Slama, K., Ray, A., Schulman, J., Hilton, J., Kelton, F., Miller, L., Simens, M., Askell, A., Welinder, P., Christiano, P., Leike, J., and Lowe, R. (2022).
\newblock Training language models to follow instructions with human feedback.

\bibitem[Paradis and Lebel, 1994]{ParadisLebel94}
Paradis, C. and Lebel, C. (1994).
\newblock Contrasts from segmental parameter settings in loanwords: core and periphery in quebec french.
\newblock In Dyck, C., editor, {\em Proceedings of the Montr{\'e}al-Ottawa-Toronto Phonology Workshop}, volume~13, pages 75--95.

\bibitem[Pater, 2005]{Pater05}
Pater, J. (2005).
\newblock Learning a stratified grammar.
\newblock In Brugos, A., Clark-Cotton, M.~R., and Ha, S., editors, {\em The Proceedings of the 29th {Boston} {University} Conference on Language Development}, pages 482--492, Somerville, MA. Cascadilla Press.

\bibitem[Pierrehumbert, 2006]{Pierrehumbert06}
Pierrehumbert, J. (2006).
\newblock The statistical basis of an unnatural alternation.
\newblock In Goldstein, L., Whalen, D.~H., and Best, C.~T., editors, {\em Laboratory phonology 8}, volume~1 of {\em {Phonology and phonetics: 4-2}}, Berlin. De Gruyter Mouton.

\bibitem[Postal, 1969]{Postal69}
Postal, P.~M. (1969).
\newblock {Mohawk} vowel doubling.
\newblock {\em International Journal of American Linguistics}, 35(4):291--298.

\bibitem[Radford et~al., 2019]{Radford+19}
Radford, A., Wu, J., Child, R., Luan, D., Amodei, D., and Sutskever, I. (2019).
\newblock Language models are unsupervised multitask learners.
\newblock Technical report, OpenAI, {S}an {F}rancisco, {CA}, {USA}.

\bibitem[Rathi et~al., 2024]{Rathi+24}
Rathi, N., Waldon, B., and Degen, J. (2024).
\newblock Informativity and accessibility in incremental production of the dative alternation.
\newblock In {\em Proceedings of the Annual Meeting of the Cognitive Science Society}, volume~46.

\bibitem[Rissanen and Ristad, 1994]{RissanenRistad94}
Rissanen, J. and Ristad, E.~S. (1994).
\newblock Language acquisition in the {MDL} framework.
\newblock In {\em Language Computations: DIMACS workshop on human language}, pages 149--166, Philedelphia. American Mathemtatical Society.

\bibitem[Rose and Walker, 2004]{RoseWalker04}
Rose, S. and Walker, R. (2004).
\newblock A typology of consonant agreement as correspondence.
\newblock {\em Language}, 80(3):475--531.

\bibitem[Roseberg and Hirschberg, 2007]{RosebergHirschberg07}
Roseberg, A. and Hirschberg, J. (2007).
\newblock V-measure: A conditional entropy-based external cluster evaluation measure.

\bibitem[Sapir and Hoijer, 1967]{SapirHoijer67}
Sapir, E. and Hoijer, H. (1967).
\newblock {\em The phonology and morphology of the {Navaho} language}.
\newblock University of California Press, Berkeley, CA.

\bibitem[Sethuraman, 1994]{Sethuraman94}
Sethuraman, J. (1994).
\newblock A constructive definition of {Dirichlet} priors.
\newblock {\em Statistica Sinica}, 4(2):639--650.

\bibitem[Shalev-Shwartz and Ben-David, 2014]{shalev-shwartz.s:2014}
Shalev-Shwartz, S. and Ben-David, S. (2014).
\newblock {\em Understanding Machine Learning: {F}rom Theory to Algorithms}.
\newblock Cambridge University Press, Cambridge, England.

\bibitem[Sinclair et~al., 2022]{Sinclair+22}
Sinclair, A., Jumelet, J., Zuidema, W., and Fern{\'a}ndez, R. (2022).
\newblock Structural persistence in language models: Priming as a window into abstract language representations.
\newblock {\em Transactions of the Association for Computational Linguistics}, 10:1031--1050.

\bibitem[Sinclair, 1987]{Sinclair87}
Sinclair, J., editor (1987).
\newblock {\em Looking up: Account of the {COBUILD} Project in Lexical Computing}.
\newblock Collins Cobuild dictionaries. Collins CoBUILD, London, England.

\bibitem[Smith, 1999]{Smith99}
Smith, J. (1999).
\newblock Noun faithfulness and accent in {Fukuoka} {Japanese}.
\newblock In Bird, S., Carnie, A., Haugen, J.~D., and Norquest, P., editors, {\em Proceedings of the 18th West Coast Conference on Formal Linguistics ({WCCFL})}, pages 519--531, Somerville, MA. Cascadilla Press.

\bibitem[Smith, 2016]{Smith16}
Smith, J. (2016).
\newblock Segmental noun/verb phonotactic differences are productive too.
\newblock {\em Proceedings of the Linguistic Society of America}, 1(0):16:1--15.

\bibitem[Stevenson and Lindberg, 2010]{NewOxfordAmericanDict3ed}
Stevenson, A. and Lindberg, C.~A., editors (2010).
\newblock {\em New {Oxford} {American} Dictionary 3rd Edition}.
\newblock Oxford University Press.

\bibitem[Teh, 2006]{Teh06}
Teh, Y.~W. (2006).
\newblock A hierarchical {Bayesian} language model based on {Pitman}-{Yor} processes.
\newblock In {\em Proceedings of the 21st International Conference on Computational Linguistics and the 44th Annual Meeting of the Association for Computational Linguistics}, ACL-44, pages 985--992, Stroudsburg, PA, USA. Association for Computational Linguistics.

\bibitem[Teh et~al., 2006]{Teh+06}
Teh, Y.~W., Jordan, M.~I., Beal, M.~J., and Blei, D.~M. (2006).
\newblock Hierarchical {Dirichlet} processes.
\newblock {\em Journal of the American Statistical Association}, 101(476):1566--1581.

\bibitem[Tenenbaum, 1999]{Tenenbaum99thesis}
Tenenbaum, J.~B. (1999).
\newblock {\em A Bayesian Framework for Concept Learning}.
\newblock PhD thesis, Massachusetts Institute of Technology.

\bibitem[Tenenbaum and Griffiths, 2001]{TenenbaumGriffiths01}
Tenenbaum, J.~B. and Griffiths, T.~L. (2001).
\newblock The rational basis of representativeness.
\newblock In {\em 23rd Annual Conference of the Cognitive Science Society}, pages 84--98.

\bibitem[Trubetzkoy, 1967]{Trubetzkoy39}
Trubetzkoy, N.~S. (1939/1967).
\newblock {\em Grundz{\"u}ge der Phonologie}.
\newblock Vandenhoeck and Ruprecht, G{\"o}ttingen.

\bibitem[Vallabha et~al., 2007]{Vallabha+07}
Vallabha, G.~K., McClelland, J.~L., Pons, F., Werker, J.~F., and Amano, S. (2007).
\newblock Unsupervised learning of vowel categories from infant-directed speech.
\newblock {\em P{N}{A}{S}}, 104(33):13273--13278.

\bibitem[Vapnik, 1998]{vapnik.v:1998}
Vapnik, V.~N. (1998).
\newblock {\em Statistical Learning Theory}.
\newblock John Wiley and Sons.

\bibitem[Vaswani et~al., 2017]{Vaswani+17_AttentionIsAllYouNeed}
Vaswani, A., Shazeer, N., Parmar, N., Uszkoreit, J., Jones, L., Gomez, A.~N., Kaiser, L.~u., and Polosukhin, I. (2017).
\newblock Attention is all you need.
\newblock In Guyon, I., Luxburg, U.~V., Bengio, S., Wallach, H., Fergus, R., Vishwanathan, S., and Garnett, R., editors, {\em Advances in Neural Information Processing Systems 30}, pages 5998--6008. Curran Associates, Inc.

\bibitem[Wainwright and Jordan, 2008a]{WainwrightJordan08book}
Wainwright, M. and Jordan, M.~I. (2008a).
\newblock {\em Graphical models, exponential families, and variational inference.}
\newblock Boston : Now Publishers, 2008.

\bibitem[Wainwright and Jordan, 2008b]{WainwrightJordan08}
Wainwright, M.~J. and Jordan, M.~I. (2008b).
\newblock Graphical models, exponential families, and variational inference.
\newblock {\em Foundations and Trends in Machine Learning}, 1(1--2):1--305.

\bibitem[Wang et~al., 2011]{Wang+11}
Wang, C., Paisley, J., and Blei, D. (2011).
\newblock Online variational inference for the hierarchical {Dirichlet} process.
\newblock In Gordon, G., Dunson, D., and Dud{\'\i}k, M., editors, {\em Proceedings of the Fourteenth International Conference on Artificial Intelligence and Statistics}, volume~15 of {\em Proceedings of Machine Learning Research}, pages 752--760, Fort Lauderdale, FL. PMLR.

\bibitem[Zimmer, 1985]{Zimmer85}
Zimmer, K. (1985).
\newblock Arabic loanwords and turkish phonological structure.
\newblock {\em International Journal of American Linguistics}, 51(4):623--625.

\bibitem[Zimmer, 1969]{Zimmer69}
Zimmer, K.~E. (1969).
\newblock Psychological correlates of some {Turkish} morpheme structure conditions.
\newblock {\em Language}, 45(2):309--321.

\end{thebibliography}
\bibliographystyle{apalike}

\appendix

\section{Hyperparameters} \label{sec:hyperparameters}
Our learning simulation of English word classes adopted exactly the same model that \citet{MoritaODonnell_sublexical_phonotactics} used for that of Japanese word clusters: the trigram model with backoff smoothing based on the hierarchical DP \citep[HDP;][]{Goldwater+06,Teh+06,Futrell+phonotactics}, including the values on hyperparameters. Specifically, the concatenation parameters of the cluster assignment DP and the backoff HDP followed the gamma prior distribution $\mathrm{Gamma}(10, 10^{-1})$ (parameterized by the shape and scale), which is the standard setting also adopted by \citet[][]{Goldwater+06}, \citet{Teh+06}, \citet{Futrell+phonotactics} etc. The top level unigram prior on the segments was uniform.

The parameters of the variational approximation of the posterior inference were also set in the same way as in \citep{MoritaODonnell_sublexical_phonotactics}. Specifically, the upper bound on the number of word clusters was set to six, and that on the number of segment-generator distributions per backoff layer was set to twice the number of possible symbols: 52 phonetic segments plus two special symbols indicating the word-initial and -final positions. We optimized the variational approximator distributions using the coordinate ascent algorithm \citep{Bishop06,BleiJordan06}, and the best approximation result among 1000 runs with random initialization was reported here. Each run was terminated either when the improvement in the approximation error (measured by the evidence lower bound, or ELBO) per iteration became less than 0.1, or when the maximum number of iterations (= 5000) was reached.

\section{Supplementary Information about the Data Format} \label{sec:celex}

As noted in 
the Data section,
we trained our clustering model on the CELEX dataset \citep{CELEX}; Specifically, the training data was extracted from the \texttt{epl.cd} file of the dataset, while filtering out lemmas whose corpus frequency---reported in the ``Cob'' column---was zero. The phonetic transcription in this dataset is coded in a special format called \emph{DISC}. DISC represents each distinct segment with a single ASCII letter, and our $n$-gram model treated each of them as a unit symbol. Specifically, we adopted the the ``PhonStrsDISC'' column of the \texttt{epl.cd} file, while the phonetic transcriptions in this paper have all been translated into IPA for better readability.

\section{Identification of the Etymological Origin} \label{sec:wiki}
The CELEX database does not provide information on the etymological origin of words. Accordingly, to evaluate the alignment of our discovered clusters with the ground-truth etymology, we made use of a subset of words whose origin was identifiable in Wikipedia.
Specifically, we considered words of
Anglo-Saxon,\footnote{\escapedURL{https://en.wikipedia.org/wiki/List_of_English_words_of_Anglo-Saxon_origin}, accessed on 5 April, 2019.}
Old Norse,\footnote{\escapedURL{https://en.wikipedia.org/wiki/List_of_English_words_of_Old_Norse_origin}, accessed on 5 April, 2019.}
Dutch,\footnote{\escapedURL{https://en.wikipedia.org/wiki/List_of_English_words_of_Dutch_origin}, accessed on 5 April, 2019.}
Latin,\footnote{\escapedURL{https://en.wikipedia.org/wiki/List_of_Latin_words_with_English_derivatives}, accessed on 5 April, 2019.}
and French origin.\footnote{\escapedURL{https://en.wikipedia.org/wiki/List_of_English_words_of_French_origin_(A-C)}, accessed on 5 April, 2019.}%
\footnote{\escapedURL{https://en.wikipedia.org/wiki/List_of_English_words_of_French_origin_(D-I)}, accessed on 5 April, 2019.}%
\footnote{\escapedURL{https://en.wikipedia.org/wiki/List_of_English_words_of_French_origin_(J-R)}, accessed on 5 April, 2019.}%
\footnote{\escapedURL{https://en.wikipedia.org/wiki/List_of_English_words_of_French_origin_(S-Z)}, accessed on 5 April, 2019.}
Words with multiple origins were included in the data only if the origins were either all Germanic (Anglo-Saxon, Old Norse, or Dutch) or Latinate (Latin or French); words reported as ``Latinates of Germanic origin''\footnote{\escapedURL{https://en.wikipedia.org/wiki/List_of_English_Latinates_of_Germanic_origin}, accessed on 5 April, 2019.} were excluded from the analysis. We also excluded words that were of ambiguous origin. The resulting data amounted to 14,172 words (consisting of 3,535 Germanics and 10,637 Latinates).

\section{Predictions of DOC-Grammaticality for Individual Verbs} \label{sec:DOC_full}

In the section titled ``Syntactic Predictions from the Clustering Results'',
we reported our model predictions regarding the DOC grammaticality of dative verbs in the form of word counts per grammaticality/etymology type $\times$ model-discovered cluster. Here, we provide more detailed results, reporting the cluster-assignment probability of each individual verb.

\begin{figure*}
    \centering
    \includegraphics[width=140mm]{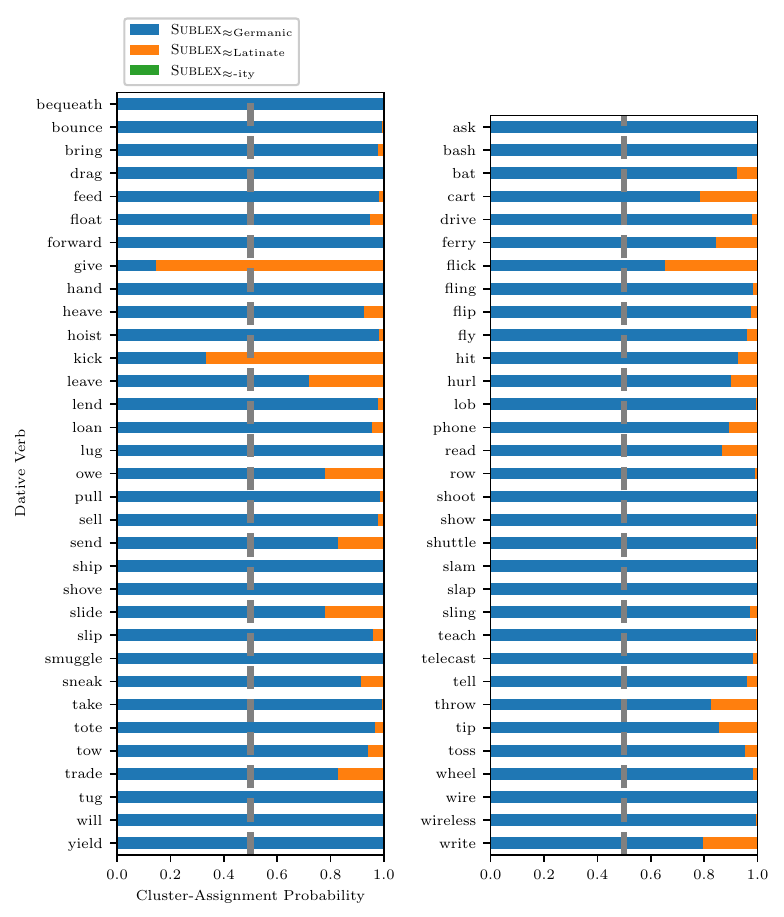}
    \caption{
        Cluster-assignment probabilities of {\supcheckMark}DOC-$\overline{\textsc{Lat}}$ dative verbs.}
    \label{fig:dative_non-Latin}
\end{figure*}

Figure~\ref{fig:dative_non-Latin} presents the posterior cluster-assignment probability of the non-Latinate dative verbs that allow DOC ({\supcheckMark}DOC-$\overline{\textsc{Lat}}$). The vertical dashed lines indicate $0.5$, which serves as an approximate decision boundary ignoring the tiny probability mass on \textsc{Sublex}\textsubscript{$\approx$-ity}.

\begin{figure*}
    \centering
    \includegraphics[width=140mm]{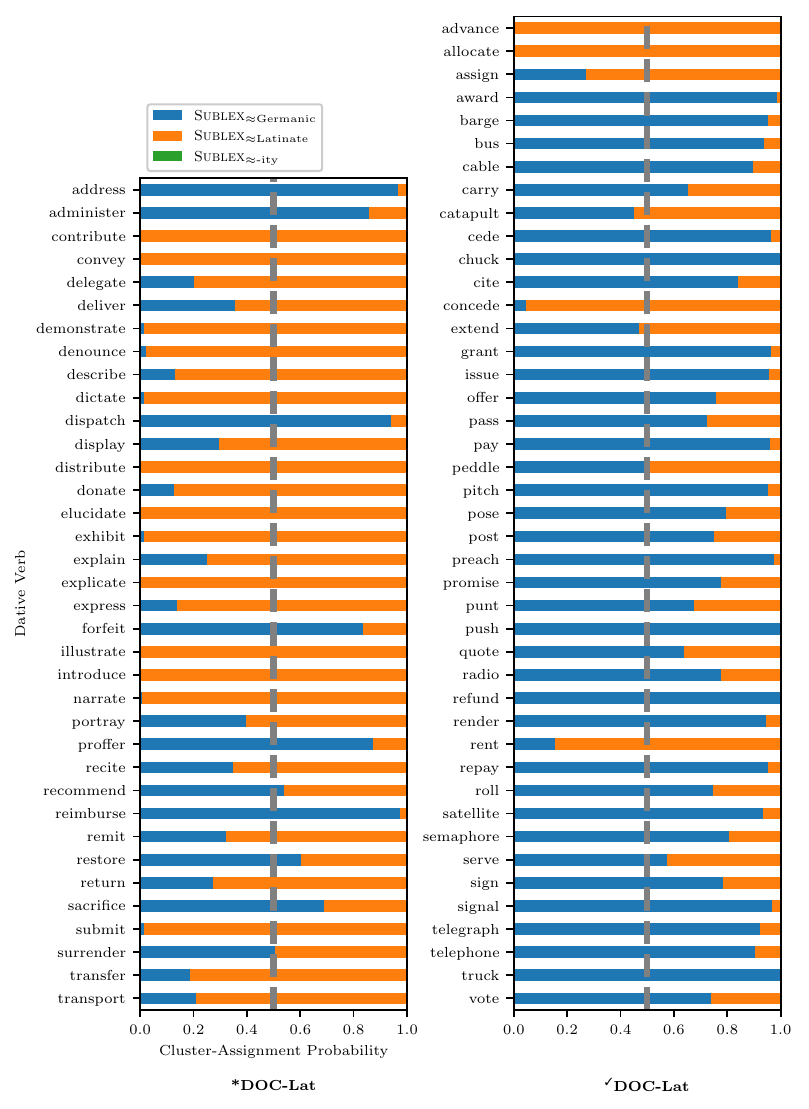}
    \caption{
        Cluster-assignment probabilities of *DOC-\textsc{Lat} (left) and {\supcheckMark}DOC-\textsc{Lat} (right) verbs.
        }
    \label{fig:dative_Latin}
\end{figure*}

We can see that the vast majority of them have a greater assignment probability to \textsc{Sublex}\textsubscript{$\approx$Germanic} (represented by the blue portions in the figure).

Latinate dative verbs prohibiting (*DOC-\textsc{Lat}) and permitting ({\supcheckMark}DOC-\textsc{Lat}) DOC are listed in Figure~\ref{fig:dative_Latin} with their cluster-assignment probabilities. Most of the *DOC-\textsc{Lat} verbs had a greater probability of assignment to \textsc{Sublex}\textsubscript{$\approx$Latinate}; by contrast, most of the {\supcheckMark}DOC-\textsc{Lat} verbs were MAP-classified into \textsc{Sublex}\textsubscript{$\approx$Germanic}. And it is these ``correct misclassifications of the exceptions'' that made our model be a better predictor of DOC grammaticality than the true etymology.

\end{document}